\documentclass[letterpaper,journal]{IEEEtran}

\usepackage{amsmath,amsfonts}
\usepackage[linesnumbered,ruled,vlined]{algorithm2e}
\usepackage{array}
\usepackage[caption=false,font=normalsize,labelfont=sf,textfont=sf]{subfig}
\usepackage{textcomp}
\usepackage{stfloats}
\usepackage{url}
\usepackage{verbatim}
\usepackage{graphicx}
\usepackage{xcolor}
\usepackage{cite}
\usepackage{verbatim}
\usepackage{amsmath}
\usepackage{amssymb}
\usepackage{algorithmicx}
\usepackage{algpseudocode}
\usepackage{booktabs}
\graphicspath{ {./fig/} }

\begin{document}

\title{Implicit Semantic-aware Communication Based on Hypergraph Reasoning}

\author{Yiwei Liao, Shurui Tu, Yong Xiao, \IEEEmembership{Senior~Member, IEEE}, Yingyu Li, Guangming~Shi, \IEEEmembership{Fellow, IEEE}

\thanks{*This work is accepted at IEEE Transactions on Communications. Copyright may be transferred without notice, after which this version may no longer be accessible.}

\thanks{This work was supported in part by the National Natural Science Foundation of China (NSFC) under grants 62525109 and 62571208, the Mobile Information Network National Science and Technology Key Project under grant 2024ZD1300700, and Hubei Natural Science Foundation Innovation Research Group Program under grant 2026AFA044. An earlier version of this paper is presented in part at the IEEE GLOBECOM Workshops, Taipei, China, December 2025\cite{tu2025hypergraph}. (Corresponding author: Yong Xiao)

Y. Liao is with China Electric Power Research Institute Co., Ltd, Wuhan 430074, China, and National Key Laboratory for Power Grid Environmental Protection, Wuhan, China. (e-mail: liaoyiwei@epri.sgcc.com.cn)

S. Tu and Y. Xiao are with the School of Electronic Information and Communications, Huazhong University of Science and Technology, Wuhan 430074, China. Y. Xiao is also with Peng Cheng Laboratory, Shenzhen, China, and Pazhou Laboratory (Huangpu), Guangzhou, China (e-mail: \{shurui\_tu, yongxiao\}@hust.edu.cn). 

Y. Li is with the School of Mechanical Engineering and Electronic Information, China University of Geosciences, Wuhan, China (e-mail: liyingyu@cug.edu.cn). 

G. Shi is with the Peng Cheng Laboratory, Shenzhen, China 518055 (e-mail: gmshi@xidian.edu.cn).}
}

\maketitle

\begin{abstract}
Semantic-aware communication has emerged as a transformative paradigm for next-generation communication systems, shifting the fundamental goal from transmitting bit-level symbols to reliably recovering and understanding the semantic meaning of information. Previous studies have demonstrated that representing the semantic content of source messages as graph-based structures can significantly improve communication efficiency and the accuracy of semantic inference at the receiver. However, existing solutions typically employ graphs that capture only pairwise relationships, thereby neglecting higher-order implicit correlations commonly observed in real-world scenarios, such as group interactions, multi-entity associations, and complex relational contexts. This limitation reduces semantic expressiveness and makes semantic inference susceptible to ambiguity and performance degradation, particularly under noisy or corrupted channel conditions. To address these issues, this paper proposes a novel hypergraph-based implicit semantic reasoning framework, called \textbf{HISR}, which leverages hypergraphs to represent complex multi-entity relationships among semantic knowledge entities. In HISR, entities and their associated higher-order relations are mapped into dedicated semantic subspaces tailored to distinct relational contexts. This design not only disentangles diverse semantic interactions to mitigate the over-smoothing effects commonly found in traditional graph embedding methods but also enables robust semantic inference even when partial information loss occurs during transmission. Extensive evaluations of benchmark datasets demonstrate that HISR consistently outperforms state-of-the-art graph-based methods, achieving substantial improvements in semantic recovery accuracy and communication robustness. These findings confirm the effectiveness and potential of hypergraph reasoning for reliable and expressive semantic communication in 6G and beyond networks. Numerical results show that the proposed HISR achieves up to a 36.6\% improvement in implicit semantic interpretation accuracy over the state-of-the-art benchmarks, confirming its ability to enhance the reliability of semantic inference while preserving the expressive advantage of hypergraph-based modeling.
\end{abstract}


\section{Introduction}

Semantic-aware communication (SAC) represents a transformative paradigm that shifts the focus of communication networking from bit-level symbol delivery to the delivery and interpretation of message meaning to the intended receivers~\cite{Shi2020semantic, XY2026SANet}. By transcending traditional transmission methods, SAC offers significant potential to improve communication efficiency, enhance Quality-of-Experience (QoE), and support emerging task-oriented applications that demand autonomous reasoning and personalized decision-making~\cite{XY2024ACMGetMobileSAN, Zhu2024SANSee}. 



Implicit semantic-aware communication (iSAC) has recently emerged as a promising solution to extend the scope of explicit semantic frameworks of SAC, which are primarily limited to the compression and transmission of directly observable semantic features~\cite{XY2024iSAC, XY2023iSAN, Deniz2024JSCCProcIEEE}. In the iSAC framework, the receiver's primary objective is to infer and reconstruct the rich implicit meaning of the source messages, including latent structures and contextual dependencies, through rigorous logical reasoning. Existing literature suggests that structured representations, particularly graph-based models, provide a robust foundation for the efficient representation, transmission, and inference of such sophisticated implicit semantics\cite{Liao_2026_WCL}. 

Despite their efficacy, existing graph-based iSAC solutions primarily focus on modeling and inferring pairwise semantic relationships. This reliance on dyadic connections inherently restricts their expressive capacity for characterizing higher-order correlations and multi-entity interactions prevalent in real-world scenarios~\cite{wen_MtransH_2016,liu_RAM_2021}. In many complex systems, such as social networks, scientific collaborations, and biological datasets, information is often structured by intricate group dependencies that go beyond simple pairwise link-based representations. Consequently, the use of pairwise models leads to an expressive bottleneck, fostering semantic ambiguity and undermining the robustness of reasoning and inference processes~\cite{contisciani_inference_2022,chien_community_2018}. 

To address these limitations, hypergraphs have emerged as a sophisticated mathematical framework that explicitly captures multi-entity and higher-order semantic relationships in a compact and expressive form. 
Unlike traditional graphs, hypergraphs use hyperedges, which represent arbitrary subsets of entities, to provide a more comprehensive characterization of complex semantic interactions~\cite{antelmi2023survey,gao_Hypergraph_Learning_2022}.


Unfortunately, the superior representational capacity of hypergraph-based representation does not automatically translate to more reliable implicit semantic reasoning. Conventional hypergraph-based inference methods typically perform high-order message aggregation within a shared embedding space. Through repeated propagation, entity embeddings may lose their individual semantic distinctions and converge toward a homogenized state, a phenomenon known as over-smoothing. Such representation homogenization is particularly harmful to the iSAC system, in which the receiver must distinguish subtle relational nuances to recover the implicit semantics not explicitly transmitted. Once these distinctions are blurred in the latent space, heterogeneous semantic relations become inseparable, degrading inference robustness—a problem exacerbated under noisy channel conditions, where fine-grained structural integrity is paramount.

Motivated by the above limitation, we propose a novel Hypergraph-based Implicit Semantic Reasoning framework, termed \textbf{HISR}, designed to enhance implicit semantic recovery by explicitly mitigating over-smoothing. Rather than conducting high-order aggregation in a singular shared space, HISR introduces a hyper-relation-based semantic subspace projection architecture. In this design, entities are projected into dedicated, relation-specific subspaces tailored to their specific semantic roles and relational contexts. This approach effectively disentangles heterogeneous interactions and preserves discriminative entity-level semantics, even under deep propagation~\cite{li2018deeper,keriven2022not}. Furthermore, to address the fundamental trade-off between expressive richness and communication efficiency, HISR incorporates an adaptive subspace configuration mechanism based on eigen-gap heuristics and subspace optimization theory. By dynamically optimizing the dimensional configurations of these relation-specific subspaces, HISR balances representational accuracy with transmission overhead, enabling high-fidelity hypergraph-based communication that is both tractable and robust to channel impairments~\cite{chen_preventing_2022}.

\begin{figure}
    \centering
  \includegraphics[width=\linewidth]{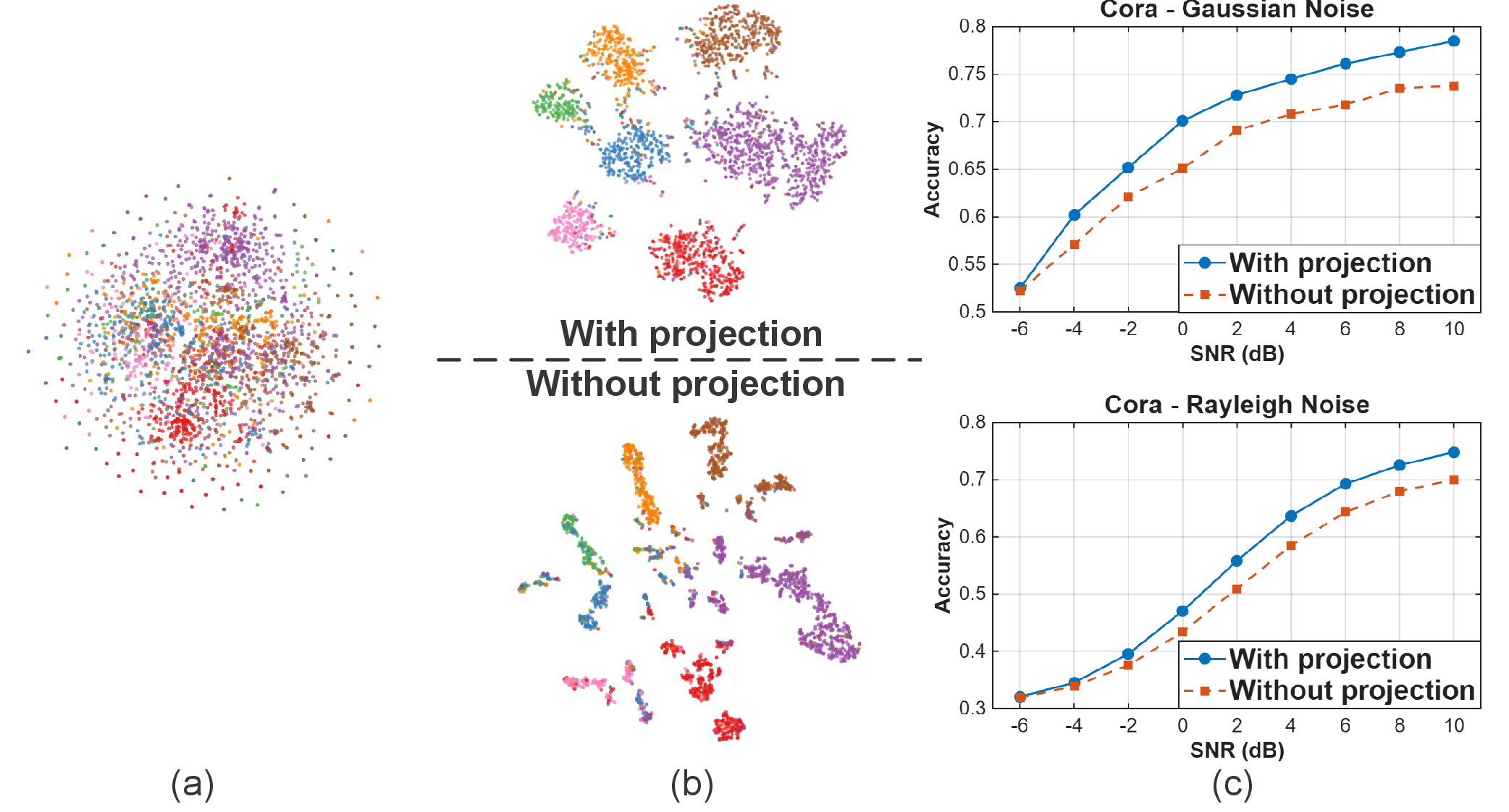}
\caption{Visualization of semantic embeddings with and without subspace projection: (a) initial embedding distribution without subspace projection, (b) embedding distribution with subspace projection, and (c) inference accuracy under Gaussian and Rayleigh noise channel on the Cora dataset. The proposed subspace projection improves semantic separability and enhances inference robustness, especially in the low-SNR regime.}
    \label{fig:intro}
\end{figure}

We illustrate the effectiveness of the proposed semantic subspace projection method in Fig.~\ref{fig:intro}. We can observe in Fig.~\ref{fig:intro}(a) that the baseline embeddings exhibit significant inter-class overlap and limited semantic separability, indicating a high degree of ambiguity. Following the application of our subspace projection framework, Fig.~\ref{fig:intro}(b) reveals that the embeddings reorganize into well-defined clusters aligned with specific relational manifolds. Furthermore, the performance curves in Fig.~\ref{fig:intro}(c) demonstrate that, compared to the traditional solutions, the refined representations achieve around 6.37\% and 6.86\% inference accuracy improvements in the Gaussian and Rayleigh fading channels, respectively.

We summarize the main contributions of this paper as follows:

\begin{itemize}
\item We propose HISR, an end-to-end iSAC framework that incorporates hypergraph modeling to explicitly represent higher-order semantic relations. By introducing relation-specific semantic subspaces, the framework enables structured encoding, transmission, and reconstruction of complex multi-entity semantics that conventional pairwise graph representations cannot capture.

\item We develop a relation-specific semantic subspace construction and inference mechanism, which disentangles heterogeneous relational contexts and alleviates the over-smoothing effect commonly observed in shared embedding spaces. This subspace-guided reasoning strategy enhances semantic discriminability and robustness under incomplete or noisy transmission conditions.

\item We establish a principled optimization framework for semantic subspace configuration, including an eigen-spectrum-based dimension selection strategy and an explicit rate--semantic trade-off formulation. 

\item Extensive experiments have been conducted based on real-world datasets to validate the performance of the proposed HISR framework. Numerical results show that the proposed HISR framework yields up to a 36.6\% improvement in implicit semantic interpretation accuracy over state-of-the-art iSAC benchmarks, which confirms its ability to enhance the reliability of semantic inference while preserving the expressive advantage of hypergraph-based modeling. 
\end{itemize}



\section{Related Work}\label{sec:related}

\begin{table*}[htbp]
    \centering
    \caption{Comparison of Graph and Hypergraph Representations in Semantic-Aware Communication}
    \label{tab:semantic_graph_comparison}
    \renewcommand{\arraystretch}{1.2}
    \begin{tabular}{p{0.14\linewidth} p{0.14\linewidth} p{0.36\linewidth} p{0.30\linewidth}}
        \toprule
        \textbf{Category} & \textbf{References} & \textbf{Strengths} & \textbf{Limitations} \tabularnewline
        \midrule
        
        \textbf{Ordinary Graph (GNN-based)} 
        & \cite{XY2023iSAN, XY2024iSAC, antelmi2023survey} 
        & Captures semantic relations by modeling entities as entities and correlations as pairwise edges. Provides effective structural encoding for semantic compression, implicit knowledge recovery, and inference tasks. 
        & Restricted to binary relations, limiting the ability to represent complex multi-entity interactions. Pairwise decomposition may lead to semantic fragmentation, especially under noisy or incomplete conditions.\tabularnewline
        \midrule
        
        \textbf{Hypergraph (HGNN-based)} 
        & \cite{feng2019hypergraph, gao2022hgnn+, keriven2022not, GETD_2020} 
        & Models higher-order relations through hyperedges that connect multiple entities simultaneously. Provides enhanced expressive power for complex semantic systems and structured knowledge representation. 
        & Shared embedding spaces may cause over-smoothing, making entity representations less distinguishable. Excessive aggregation can introduce semantic ambiguity, particularly under lossy transmission scenarios.\tabularnewline
        \bottomrule
    \end{tabular}
\end{table*}

\noindent
\textbf{Graph-Based Semantic Embedding:} Implicit semantic-aware communication (iSAC) methods have increasingly adopted structured graph representations as a fundamental tool for encoding and decoding semantic content within transmitted messages. Most existing studies utilize graph neural networks (GNNs) to capture semantic relationships by representing entities as nodes and their pairwise correlations as edges~\cite{XY2023iSAN, XY2024iSAC}. These graph-based approaches have shown substantial effectiveness in tasks such as semantic compression, implicit knowledge recovery, and robust inference. However, their reliance on pairwise relational structures inherently restricts their ability to model complex, higher-order interactions that are prevalent in real-world scenarios, such as social networks, collaborative projects, and biological systems. Consequently, the semantic integrity captured by these methods is often limited, leading to potential semantic fragmentation and degraded inference performance, especially when handling noisy or incomplete data~\cite{antelmi2023survey,GETD_2020}.

\noindent
\textbf{Hypergraph Representation and Inference:} Hypergraphs extend the capability of traditional graphs by enabling hyperedges to connect arbitrary sets of entities, thus naturally modeling higher-order, multi-entity interactions. Recent developments in HGNNs have demonstrated significant promise across various applications, including link prediction, entity classification, and clustering~\cite{feng2019hypergraph,chami2019hyperbolic,gao2022hgnn+}. Despite their enhanced representational power, existing hypergraph-based methods typically embed all entities and hyper-relations within a single shared embedding space. This uniform treatment often results in the over-smoothing phenomenon, wherein entity representations progressively become indistinguishable due to excessive information aggregation~\cite{keriven2022not,GETD_2020}. Such semantic ambiguity critically undermines the effectiveness and interpretability of hypergraph inference, particularly under challenging operational conditions such as noisy transmission channels or partial information loss.

\noindent
\textbf{Relation-Specific Subspaces:} To address the semantic ambiguity arising from uniform embeddings, recent advances in neural-symbolic reasoning advocate the use of relation-specific subspace embeddings\cite{liu_RAM_2021}. These approaches propose embedding entities into distinct subspaces according to their relational contexts, effectively decoupling heterogeneous semantic interactions~\cite{rossi2021knowledge,wang2014knowledge,shao2024theory,schlichtkrull2018modeling,Liao_2026_WCL,Liao_2026_HIB}. Relation-specific subspaces enhance the precision and robustness of semantic inference, preserving the granularity and integrity of semantic content even under adverse conditions. By maintaining clear semantic distinctions across different relations, this approach significantly reduces over-smoothing and improves inference reliability, particularly in scenarios involving incomplete, noisy, or dynamically evolving data.

\noindent
\textbf{Our Contribution:} Motivated by the aforementioned insights and limitations, we propose \textbf{HISR}, a novel hypergraph-based semantic communication framework explicitly designed to leverage relation-specific subspace embeddings. HISR systematically integrates the expressive capability of hypergraphs with advanced embedding techniques, enabling precise modeling and robust inference of complex multi-entity semantic relations. By disentangling heterogeneous semantic contexts into dedicated subspaces, HISR effectively mitigates the over-smoothing problem inherent in traditional GNN and HGNN models, thereby substantially enhancing semantic compression efficiency, inference accuracy, and robustness under lossy and noisy transmission conditions. Comprehensive experimental evaluations conducted on benchmark datasets validate the superior performance of HISR, highlighting its potential to serve as a foundational architecture for reliable, expressive, and efficient semantic-aware communication systems.

\section{System Model}

\begin{figure*}[ht]
    \centering
    \includegraphics[width=0.99\textwidth]{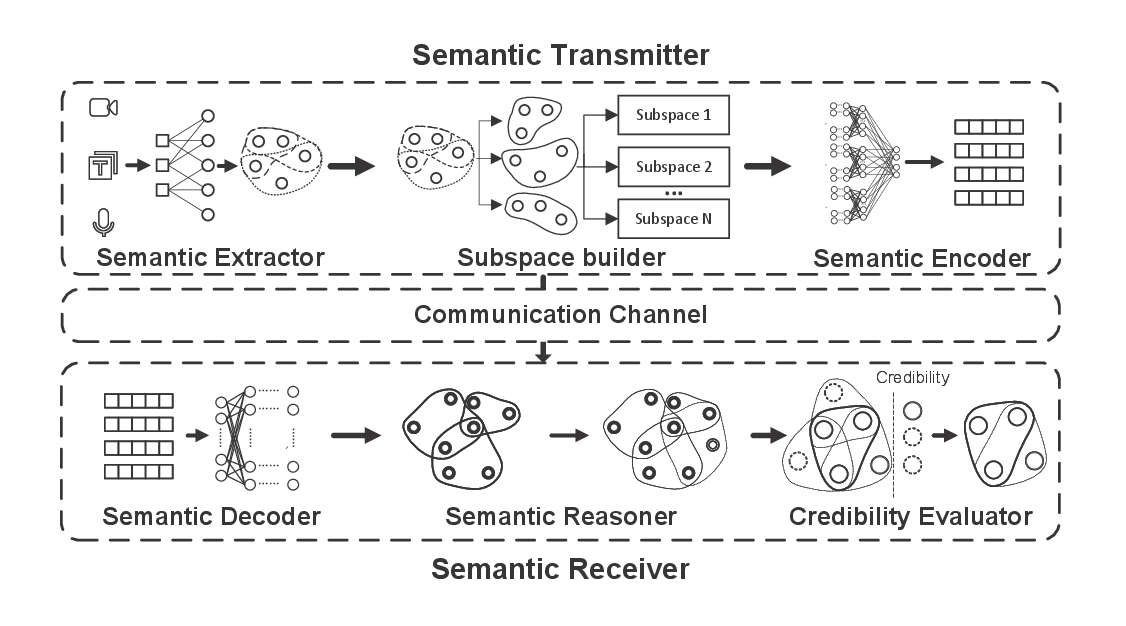}
    \caption{The proposed HISR architecture: Semantic Extraction, Relation-Specific Subspace Modeling, Knowledge Base and Implicit Semantic Reasoning for Semantic-Aware Communication}
    \label{fig:system_model}
\end{figure*}

The proposed HISR, depicted in Fig.~\ref{fig:system_model}, primarily comprises three main components: the \emph{Semantic Transmitter}, the \emph{Communication Channel}, and the \emph{Semantic Receiver}. The Semantic Transmitter consists of a semantic extractor, a semantic subspace builder, and a semantic encoder. Its role is to transform raw multimodal data into compact semantic embeddings, which are then transmitted through the communication channel. The channel typically introduces noise and distortions, mimicking realistic communication scenarios. At the receiving end, the Semantic Receiver, composed of a semantic decoder, reasoner, and evaluator, reconstructs and infers the implicit multi-entity semantic content embedded within the transmitted signal, ensuring robust and accurate recovery of complex semantic relationships.

\subsection{Semantic Representation and Semantic Extractor}

We model the underlying semantic knowledge as a structured hypergraph, formally denoted by \(\mathcal{H} = (\mathcal{V}, \mathcal{E}, \mathcal{R})\), explicitly designed to preserve and encode the rich semantic content inherent in multi-entity interactions. In this formulation, the set \(\mathcal{V}\) is the set of semantic entities, each of which corresponds to a distinct semantic concept, such as an individual object, an agent, an event, or an abstract notion. The set \(\mathcal{E}\) of hyperedges represents complex semantic relationships, explicitly capturing interactions involving multiple entities simultaneously. Such higher-order relations are essential for accurately representing semantic phenomena beyond simple binary relations, reflecting the true complexity encountered in real-world knowledge and communication scenarios.

Each hyperedge \( e \in \mathcal{E} \) is formally defined as a tuple \( e = (v_1, v_2, \dots, v_{k_r}; r) \), where the entities \(\{v_i\}_{i=1}^{k_r}\) are associated through a specific semantic relation type \( r \in \mathcal{R} \). The relation type \( r \) further characterizes the nature and contextual semantics of these entity interactions, providing essential semantic clarity. For instance, in a knowledge hypergraph, a relation might denote a collaborative event, a causal interaction, or a shared attribute among multiple entities. Each relation type \( r \) is associated with a specific arity \( k_r \), indicating precisely how many entities are involved in that relation. This detailed specification allows the hypergraph to maintain the semantic coherence of high-order interactions, ensuring the precise semantic role of each entity within complex semantic structures.

The explicit representation of these multi-entity semantic relations within a hypergraph structure provides several advantages for semantic-aware communication. First, it naturally captures semantic richness, avoiding the limitations of decomposing complex interactions into simpler pairwise relations, which often leads to loss of critical contextual information. Second, the relational context explicitly encoded by hyperedges greatly enhances the semantic precision and reduces ambiguity, providing a more robust foundation for downstream semantic inference tasks.

The Semantic Extractor plays a pivotal role in our hypergraph framework, transforming multimodal and heterogeneous raw input data (e.g., textual, visual, audio) into structured semantic representations. Specifically, it extracts semantic entities and relations, embedding each semantic entity into an initial structured vector \(\mathbf{x}_v \in \mathbb{R}^{d}\), where \(d\) is carefully selected to preserve semantic distinctions and support efficient transmission. These embeddings serve as essential inputs for subsequent semantic subspace construction and inference, effectively maintaining semantic coherence and relational consistency throughout the communication pipeline, thereby significantly enhancing robustness and precision in implicit semantic communication.

\subsection{Semantic Subspace Builder}

To accurately and robustly represent distinct semantic contexts within complex multi-entity interactions, we introduce a specialized semantic subspace embedding approach. Unlike traditional embedding strategies that project all entities into a unified latent space, which often results in semantic ambiguity, our method explicitly constructs dedicated semantic subspaces tailored to distinct relation types. By doing so, each semantic subspace precisely encapsulates unique relational semantics, ensuring clear boundaries between different semantic contexts and preserving semantic coherence within each subspace.

Formally, let \(\mathbf{z}_v^{(l)} \in \mathbb{R}^{d}\) denote the embedding of entity \( v \) at layer \( l \). To construct effective semantic embeddings that respect relation-specific semantics, we introduce an iterative embedding update mechanism governed by:
\begin{equation}
\mathbf{z}_v^{(l+1)} = \sigma\left(\sum_{e \in \mathcal{E}(v)} \frac{1}{|\mathcal{N}_e|}\sum_{u \in \mathcal{N}_e}\mathbf{z}_u^{(l)}W_e + \mathbf{C}_v\right),
\end{equation}
where \(\mathcal{E}(v)\) represents the set of hyperedges containing entity \( v \), and \(\mathcal{N}_e\) is the entity set within hyperedge \( e \). Here, \(W_e \in \mathbb{R}^{d\times d}\) denotes a hyperedge-specific transformation matrix, encoding the distinct semantic characteristics of hyperedge \( e \). The vector \(\mathbf{C}_v \in \mathbb{R}^{d}\) serves as a local correction term designed explicitly to preserve and enhance the semantic identity of entity \( v \), thus preventing semantic drift and ensuring robust semantic coherence. The non-linear activation \(\sigma(\cdot)\), such as ReLU, further promotes effective semantic aggregation by introducing necessary non-linearities.

To ensure precise semantic distinctions, we further introduce a mechanism for selecting and refining relation-specific subspaces based on semantic consistency within hyperedges. For each hyperedge \( e \in \mathcal{E} \), the semantic coherence within the subspace is quantitatively evaluated by computing pairwise semantic discrepancies:
\begin{equation}\label{eq:semantic_coherence}
(u_e, v_e)=\arg\max_{u,v\in e}\|\mathbf{F}_{u\triangleleft e}\mathbf{z}_u - \mathbf{F}_{v\triangleleft e}\mathbf{z}_v\|_2,
\end{equation}
where the linear transformation \(\mathbf{F}_{u\triangleleft e}\in\mathbb{R}^{d\times d}\) maps entity embeddings into hyperedge-specific subspaces.

Subsequently, we leverage these selected discrepant pairs to build an auxiliary graph structure \( G_H=(\mathcal{V},E') \), linking entities within the same hyperedge only through the pair of entities exhibiting the highest semantic difference. This targeted connectivity significantly simplifies semantic relationships and preserves computational efficiency without sacrificing semantic expressivity.

Moreover, the resulting semantic embeddings are optimized through a relation-specific Laplacian-based regularization strategy, defined as follows:
\begin{equation}\label{eq:semantic_subspace_regularization}
\mathbf{z}_v^{(l+1)} = \mathbf{z}_v^{(l)} - \eta\,\mathcal{L}_{\mathrm{NL}}(\mathbf{z}_v^{(l)}),
\end{equation}
with the non-linear Laplacian operator \(\mathcal{L}_{\mathrm{NL}}\) given by:
\begin{equation}
(\mathcal{L}_{\mathrm{NL}}\mathbf{z})_v = \sum_{e;u_e\sim_e v}\frac{1}{\delta_e}\mathbf{F}_{v\triangleleft e}^{\top}\left(\mathbf{F}_{v\triangleleft e}\mathbf{z}_v-\mathbf{F}_{u_e\triangleleft e}\mathbf{z}_{u_e}\right),
\end{equation}
where \(\delta_e\) normalizes contributions from hyperedge \(e\), and \(\eta\) is the learning rate. This iterative refinement strategy explicitly promotes the convergence of embeddings within each semantic subspace, ensuring tight semantic coherence and effectively preventing the common over-smoothing issue observed in traditional embedding frameworks.

The explicit subspace construction strategy further allows for precise modeling of semantic interactions and inference scenarios. By maintaining clear semantic boundaries between different semantic contexts, each subspace facilitates robust and accurate semantic reasoning, even when faced with incomplete or noisy information during transmission, our proposed Semantic Subspace Builder rigorously preserves and captures the essential semantic information of complex higher-order interactions, providing a robust and semantically precise foundation for subsequent semantic communication tasks.

\subsection{Semantic Encoder}

Following the semantic subspace embedding process, the next critical step is to effectively aggregate semantic information from individual entity embeddings into structured hyperedge representations, ensuring robust semantic communication. In particular, semantic relations typically involve interactions among multiple entities, and capturing these interactions precisely requires an aggregation mechanism that explicitly respects and leverages the distinct semantic contexts defined by relation-specific subspaces.

Given a hyperedge \( e = (v_1, v_2, \dots, v_k; r) \), where each entity \( v_i \) is represented by a semantic embedding \( \mathbf{z}_{v_i}^{(r)} \in \mathbb{R}^{d} \) within the corresponding semantic subspace associated with relation \( r \), we define a relation-specific semantic encoder to aggregate and encode these embeddings into a unified hyperedge representation \(\mathbf{h}_e\). This aggregated representation serves as the semantic carrier, compactly encapsulating the combined semantic content of all constituent entities, enabling efficient and robust semantic transmission over the communication channel.

Formally, the semantic hyperedge embedding \(\mathbf{h}_e\) is computed using a dedicated aggregation function \(\phi_r(\cdot)\), designed specifically for relation \(r\):
\begin{equation}
\mathbf{h}_e = \phi_r\left(\{\mathbf{z}_{v_i}^{(r)}\}_{i=1}^{k}\right) 
= \sigma\left(W_r\,\frac{1}{k}\sum_{i=1}^{k}\mathbf{z}_{v_i}^{(r)} + \mathbf{b}_r'\right),
\end{equation}
where the transformation matrix \( W_r \in \mathbb{R}^{d \times d} \) and bias vector \( \mathbf{b}_r' \in \mathbb{R}^{d} \) are learnable parameters specifically tailored to relation type \( r \). These parameters allow the encoder to explicitly capture and highlight relation-specific semantic characteristics, effectively encoding unique relational semantics into hyperedge embeddings. The non-linear activation function \( \sigma(\cdot) \), typically a ReLU or similar activation function, introduces essential non-linearity, enabling the model to capture complex semantic interactions and relationships within each hyperedge context.

The encoder aggregates entity embeddings within each hyperedge by averaging \(\frac{1}{k}\sum_{i=1}^{k}\mathbf{z}_{v_i}^{(r)}\) to ensure balanced contributions and semantic coherence, while the relation-specific transformation \(W_r\) selectively emphasizes dimensions relevant to each relational context. The resulting hyperedge embedding \(\mathbf{h}_e\) provides a compact, relation-aware representation that reduces communication complexity and enables robust semantic recovery over noisy channels, simultaneously enhancing precision and reducing ambiguity in downstream inference tasks.

\subsection{Semantic Receiver}

In practical semantic communication scenarios, transmitted semantic embeddings inevitably suffer from channel-induced distortion. Let $\mathbf{h}_e$ denote the transmitted embedding for hyperedge $e$, and the receiver observes
\begin{equation}
\hat{\mathbf{h}}_e = \mathbf{h}_e + \mathbf{n}_e,
\end{equation}
where $\mathbf{n}_e$ represents channel noise. This perturbation introduces semantic uncertainty, potentially shifting the embedding away from its intended relational subspace.

Upon receiving the transmitted hyperedge embeddings \(\{\hat{\mathbf{h}}_e\}\), the Semantic Receiver performs a structured decoding and inference procedure to accurately reconstruct and reason about the original semantic information. Specifically, this process leverages the relation-specific semantic subspaces explicitly constructed in the encoding phase, aiming to effectively handle partial information loss, channel-induced distortions, and ambiguities inherent in semantic communication scenarios.

\subsubsection{Subspace-Aware Decoding}

At the semantic receiver, the first critical step involves decoding the received hyperedge embeddings back into their corresponding semantic subspaces. Each received hyperedge embedding \(\hat{\mathbf{h}}_e\) associated with a specific relation type \( r \) is projected into the corresponding semantic subspace, facilitating accurate identification of the candidate entity tuples. This ensures that semantic decoding strictly adheres to the relation-specific semantic boundaries established during encoding. Formally, given the set of candidate entity embeddings \(\{\mathbf{z}_{v_i}^{(r)}\}\), the decoding procedure generates a set of candidate entity tuples defined by:
\begin{equation}
\mathcal{C}_r(e) = \left\{(v_1, v_2, \dots, v_k): \|\hat{\mathbf{h}}_e - \phi_r(\{\mathbf{z}_{v_i}^{(r)}\})\|\leq\tau_r\right\},
\end{equation}
where \(\tau_r\) is a semantic coherence threshold determined through validation, optimized to maintain semantic proximity and consistency within each relation-specific semantic subspace.

\subsubsection{Cross-subspace projection and Completion}

Given the inherent complexity of semantic interactions and the potential ambiguity arising from incomplete or corrupted information, robust semantic inference necessitates effectively leveraging evidence across multiple semantic subspaces. To this end, the semantic receiver introduces a cross-subspace projection strategy to comprehensively aggregate and evaluate semantic coherence across various relation-specific subspaces.

We define a composite semantic confidence score \(S(e)\) to quantify semantic coherence across candidate tuples from multiple subspaces:
\begin{equation}
S(e) = \sum_{r\in \mathcal{R}} \beta_r \max_{(v_1,\dots,v_k)\in \mathcal{C}_r(e)} f_r\left(\hat{\mathbf{z}}_{v_1}^{(r)},\hat{\mathbf{z}}_{v_2}^{(r)},\dots,\hat{\mathbf{z}}_{v_k}^{(r)}\right),
\end{equation}
where \(f_r(\cdot)\) is a trainable neural scoring function designed to evaluate semantic consistency within the subspace corresponding to relation \( r \). The normalization weights \(\beta_r\) effectively balance the semantic contributions from different subspaces, allowing adaptive inference that selectively emphasizes reliable subspace information and mitigates semantic uncertainty caused by noisy channels.

The scoring function \(f_r(\cdot)\) is further defined by leveraging a neural network-based positional-aware mechanism:
\begin{equation}
f_r(\hat{\mathbf{z}}_{v_1}^{(r)},\dots,\hat{\mathbf{z}}_{v_k}^{(r)})=\frac{1}{k}\sum_{p=1}^{k}f_r^{(p)}(\hat{\mathbf{z}}_{v_1}^{(r)},\dots,\hat{\mathbf{z}}_{v_k}^{(r)}),
\end{equation}
where the individual positional scores \(f_r^{(p)}\) measure the semantic coherence considering specific entity positions and subspace interactions, thus accurately reflecting complex positional semantics.

\subsubsection{Hyper-Relation Reconstruction}

Based on the computed composite semantic confidence scores, the semantic receiver reconstructs the semantic hyperedge set by selecting hyperedges whose confidence surpasses a predefined global threshold \(\gamma\):
\begin{equation}
\hat{\mathcal{E}} = \{ e : S(e) \geq \gamma \},
\end{equation}

The threshold \(\gamma\) is empirically tuned to optimize precision-recall trade-offs, ensuring robust and reliable semantic hyperedge reconstruction. Finally, the reconstructed hypergraph \(\hat{\mathcal{H}}=(\hat{\mathcal{V}},\hat{\mathcal{E}},\mathcal{R})\) is evaluated against the original ground-truth hypergraph to rigorously assess semantic inference performance, validating the semantic accuracy, precision, and robustness of the inference pipeline.

In practical deployment scenarios, relational labels may be incomplete or corrupted upon reception due to noisy transmission channels. To ensure robustness, the semantic receiver incorporates a multi-hypothesis inference strategy across semantic subspaces. Specifically, when the relation label of a received embedding is uncertain, the receiver performs scoring across multiple candidate relation-specific subspaces simultaneously, selecting the most semantically consistent hypothesis. Formally, the semantic credibility score in such scenarios is computed as:
\begin{equation}
    S(e) = \max_{r \in \mathcal{R}} \beta_r f_r(\hat{z}_e),
\end{equation}
where $\beta_r$ denotes the reliability weighting factor reflecting prior confidence or decoding uncertainty for relation $r$. This multi-hypothesis strategy effectively compensates for incomplete or corrupted relational information, enhancing robustness of the implicit semantic reconstruction.

\subsubsection{Complexity and Robustness Analysis of the Receiver}

The computational complexity at the semantic receiver primarily involves projecting received embeddings back into semantic subspaces, candidate tuple identification, and evaluating semantic coherence scores. Given \( |\mathcal{E}| \) hyperedges, \( |\mathcal{R}| \) relation types, and embedding dimension \( d \), the complexity of the decoding and inference procedures is approximately:
\begin{equation}
\mathcal{O}\left(|\mathcal{E}|\cdot|\mathcal{R}|\cdot d^2\right),
\end{equation}
highlighting the practical scalability of our method. Additionally, by explicitly aggregating semantic information across multiple subspaces and employing robust positional scoring mechanisms, the semantic receiver demonstrates superior resilience and semantic precision under realistic noisy and incomplete channel conditions.

Overall, the structured semantic inference approach at the receiver side systematically exploits semantic subspace structures and positional semantics, significantly enhancing semantic reconstruction accuracy and robustness, making it highly effective for real-world implicit semantic-aware communication scenarios.

\subsection{Knowledge Base}

The semantic transmitter only transmits compact semantic embeddings obtained through the semantic extractor, subspace builder, and semantic encoder. The semantic receiver maintains a pre-shared hypergraph knowledge base $\mathcal{H}_{\text{KB}}$, which is constructed offline from historical data sources. Thus, the receiver utilizes this pre-shared semantic hypergraph to reconstruct and infer implicit semantics that are not explicitly transmitted. Consequently, the communication task explicitly focuses on transmitting minimal yet essential semantic information, relying on the receiver-side knowledge base for comprehensive semantic reasoning and reconstruction.

The knowledge base (KB) serves as a persistent semantic memory layer that supports implicit semantic inference at the receiver. Formally, we model the KB as a hypergraph

\begin{equation}
\mathcal{H}_{\text{KB}} = (\mathcal{V}, \mathcal{E}_{\text{KB}}, \mathcal{R}),
\end{equation}

where $\mathcal{V}$ denotes the global entity set, $\mathcal{E}_{\text{KB}}$ represents the stored hyper-relational tuples, and $\mathcal{R}$ is the relation type set consistent with the semantic subspace construction.

The KB is initialized through an offline construction phase. Given historical semantic observations $\mathcal{D} = \{\tau_i\}$, where each tuple $\tau_i = (v_1, \dots, v_k; r)$ corresponds to a $k$-ary relation, the hyperedge set is formed as

\begin{equation}
\mathcal{E}_{\text{KB}} = \bigcup_i \tau_i,
\end{equation}

In practice, these tuples can be extracted from structured databases, textual corpora, or domain-specific event streams via lightweight entity recognition and relation extraction methods. Importantly, higher-order relations are stored directly in hyperedge form without decomposition into pairwise edges, thereby preserving semantic completeness.

During system operation, the KB evolves dynamically as new observations arrive in the form of $\Delta \mathcal{D}_t$ at time $t$. Rather than reconstructing the entire structure, incremental updates are performed as

\begin{equation}
\mathcal{E}_{\text{KB}}^{(t+1)} 
= 
\mathcal{E}_{\text{KB}}^{(t)} 
\cup 
\Delta \mathcal{E}^{(t)},
\end{equation}

where $\Delta \mathcal{E}^{(t)}$ contains newly observed or modified hyper-relational tuples. Each update is associated with a version index $\nu^{(t)}$ and temporal metadata, enabling efficient synchronization while keeping communication overhead bounded. 

In distributed environments, multiple receiver entities may maintain local replicas $\mathcal{H}_{\text{KB}}^{(i)}$. Consistency is maintained through periodic synchronization of differential tuples and version control, with conflicts resolved via timestamp-based ordering or lightweight consensus mechanisms. Since the proposed semantic inference operates within relation-specific subspaces and tolerates partial knowledge, strict instantaneous consistency is not required; eventual consistency is sufficient to ensure reliable reasoning performance.

\section{Semantic Subspace Construction}\label{sec:semantic_subspace}

In this section, we introduce and analyze the \emph{Semantic Subspace Construction} component of HISR, which effectively captures higher-order semantic interactions and provides robust foundations for semantic reasoning and inference.

\begin{figure}
    \centering
    \includegraphics[width=1\linewidth]{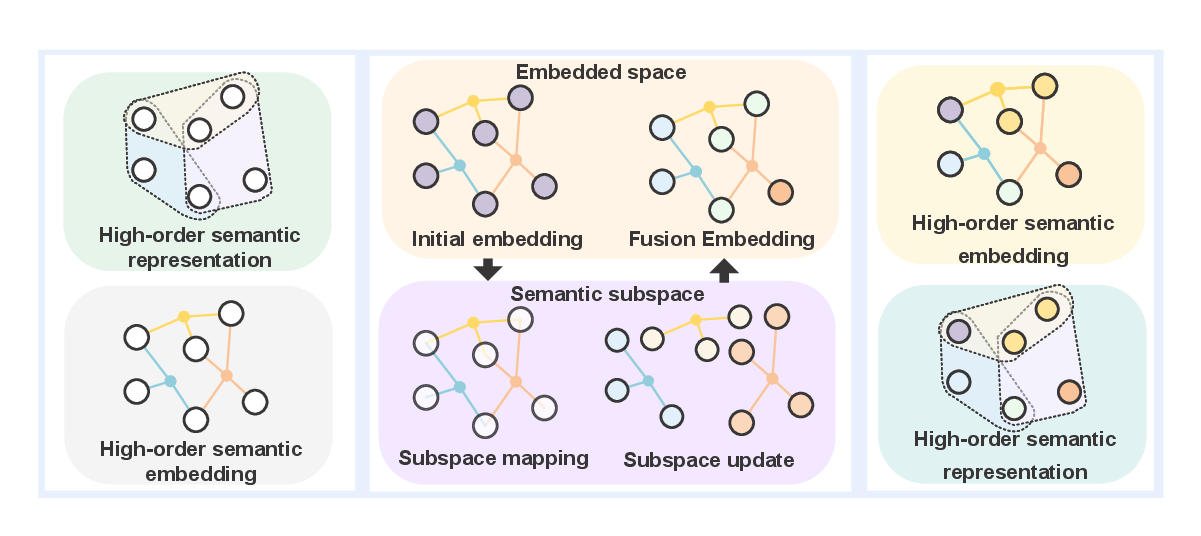}
    \caption{Information completion based on relation-specific semantic subspaces}
    \label{embedding subspace}
\end{figure}

\subsection{Building process}

Hypergraphs are typically defined by higher-order interactions among multiple entities. While traditional hypergraph Laplacians can aggregate entities into homogeneous representations, this often results in semantic ambiguity and the notorious over-smoothing problem. To mitigate these limitations, we introduce semantic subspace theory to enrich the hypergraph structure with additional geometric semantics.

Formally, a semantic subspace $\mathcal{F}$ on a hypergraph $\mathcal{H}=(\mathcal{V},\mathcal{E})$ is defined by associating to each entity $v \in \mathcal{V}$ a vector space called the vertex stalk $\mathcal{F}(v)$, and to each hyperedge $e \in \mathcal{E}$ a vector space called the hyperedge stalk $\mathcal{F}(e)$. For each incidence relation $(v \triangleleft e)$ indicating entity $v$ belongs to hyperedge $e$, we define a linear map known as the restriction map:
\begin{equation}
    \mathbf{F}_{v\triangleleft e}:\mathcal{F}(v)\rightarrow\mathcal{F}(e),
\end{equation}

which encodes how information propagates from entities to hyperedges.

Fig.~\ref{embedding subspace} illustrates the core principle of our proposed relation-specific semantic subspace construction. Unlike traditional embedding methods that place all semantic entities into a unified embedding space—often resulting in ambiguity and loss of higher-order relational distinctions—our approach explicitly partitions entities into distinct, relation-specific semantic subspaces. As shown at the top, conventional embedding methods aggregate entities related by different semantic relations ($e_1$, $e_2$, and $e_3$) into a single space, causing overlap and potential semantic confusion. In contrast, our method (middle of the figure) clearly separates these entities into separate subspaces (Subspace 1, Subspace 2, and Subspace 3), each corresponding uniquely to a specific semantic relation. The resulting embeddings (bottom of the figure) demonstrate clear and distinct boundaries between different semantic relations, significantly enhancing semantic clarity and robustness. This explicit subspace-based modeling ensures accurate semantic inference and mitigates the over-smoothing effect common in traditional hypergraph embedding frameworks.

Building upon these subspace definitions, we introduce two variants of the subspace hypergraph Laplacian: linear and non-linear. In this paper, we focus on the more expressive non-linear variant due to its robustness and semantic clarity. Specifically, for each hyperedge $e$, we first identify the most semantically discrepant pair of entities by:
\begin{equation}\label{eq:max_discrepancy}
(u_e,v_e)=\arg\max_{u,v\in e}\|\mathbf{F}_{u\triangleleft e}\mathbf{z}_u-\mathbf{F}_{v\triangleleft e}\mathbf{z}_v\|_2 ,
\end{equation}
where $\mathbf{z}_u$ denotes the embedding of entity $u$. Using these pairs, we construct a simplified graph structure $G_H=(\mathcal{V}, E')$ that significantly reduces computational complexity compared to traditional clique-based constructions.

The corresponding non-linear subspace hypergraph Laplacian operator $\mathcal{L}_{\mathrm{NL}}$ is defined by:
\begin{equation}
(\mathcal{L}_{\mathrm{NL}}\mathbf{z})_v=\sum_{e;u_e\sim_e v}\frac{1}{\delta_e}\mathbf{F}_{v\triangleleft e}^{\top}\left(\mathbf{F}_{v\triangleleft e}\mathbf{z}_v-\mathbf{F}_{u_e\triangleleft e}\mathbf{z}_{u_e}\right),
\end{equation}
where $\delta_e$ normalizes contributions from each hyperedge $e$.

This non-linear operator induces embeddings to converge within distinct semantic subspaces, minimizing the generalized subspace total variation (TV) energy function:

\begin{small}
\begin{equation}
E_{\mathrm{TV}}(\mathbf{z})=\frac{1}{2}\sum_{e\in\mathcal{E}}\frac{1}{\delta_e}\max_{u,v\in e}\|\mathbf{F}_{v\triangleleft e}\mathbf{D}_v^{-\frac{1}{2}}\mathbf{z}_v-\mathbf{F}_{u\triangleleft e}\mathbf{D}_u^{-\frac{1}{2}}\mathbf{z}_u\|_2^2,
\end{equation}
\end{small}

where $\mathbf{D}_v=\sum_{e:v\in e}\mathbf{F}_{v\triangleleft e}^{\top}\mathbf{F}_{v\triangleleft e}$ is a normalization analogous to entity degree.




\begin{algorithm}[ht]
\caption{Semantic Subspace Construction}
\label{alg:subspace_subspace}
\LinesNotNumbered
\KwIn{Hypergraph $\mathcal{H}=(\mathcal{V},\mathcal{E})$, initial embeddings $\{\mathbf{z}_v^{(0)}\}_{v\in\mathcal{V}}$, maximum iterations $L$, convergence threshold $\epsilon$}
\KwOut{Relation-specific semantic embeddings $\{\mathbf{z}_v^{(L)}\}_{v\in\mathcal{V}}$}

\For{$l=0$ \KwTo $L-1$}{
    \For{each entity $v\in\mathcal{V}$}{
        Initialize message aggregation $\mathbf{m}_v^{(l)}=\mathbf{0}$\;
        \For{each hyperedge $e\in\mathcal{E}(v)$ containing entity $v$}{
            Compute most discrepant pair $(u_e,v_e)$ via Eq.~\eqref{eq:max_discrepancy}\;
            \If{$v=v_e$}{
                Update aggregation:
                \[
                \mathbf{m}_v^{(l)}\gets\mathbf{m}_v^{(l)}+\frac{1}{\delta_e}\mathbf{F}_{v\triangleleft e}^{\top}(\mathbf{F}_{v\triangleleft e}\mathbf{z}_v^{(l)}-\mathbf{F}_{u_e\triangleleft e}\mathbf{z}_{u_e}^{(l)})
                \]
            }
        }
        Update embedding with non-linear diffusion \tcp*{$\eta$ is the learning rate, $\sigma$ is ReLU}
        \[
        \mathbf{z}_v^{(l+1)}\gets\sigma(\mathbf{z}_v^{(l)}-\eta\,\mathbf{m}_v^{(l)})
        \]
    }
    \If{$\frac{\|\mathbf{Z}^{(l+1)}-\mathbf{Z}^{(l)}\|_F}{\|\mathbf{Z}^{(l)}\|_F}<\epsilon$}{
        break\tcp*[f]{Stopping criterion met}
    }
}
\Return{Final semantic embeddings $\{\mathbf{z}_v^{(L)}\}_{v\in\mathcal{V}}$\;}

\end{algorithm}

Additionally, the learning rate $\eta$ in the iterative update plays a critical role in determining convergence stability and the quality of the resulting semantic embeddings. A properly chosen learning rate ensures stable diffusion dynamics and effective minimization of the subspace energy. If $\eta$ is excessively large, the update process may become unstable and fail to converge; conversely, if $\eta$ is too small, convergence may be significantly slowed, leading to inefficient optimization. In practice, $\eta$ should be selected to balance convergence speed and stability, thereby ensuring reliable semantic subspace construction and robust inference performance under practical deployment conditions.

\subsection{Optimization objectives and complexity} 

To avoid ambiguity, we clarify that each relation type $r \in \mathcal{R}$ corresponds to one dedicated semantic subspace. Therefore, the total number of semantic subspaces equals $|\mathcal{R}|$, while the following equations determine the optimal dimension $d_r$ of each subspace.

\subsubsection{Optimal subspace dimension} 

Let $Z \in \mathbb{R}^{n\times d}$ denote the semantic embedding matrix, where $n$ is the number of entities and $d$ is the embedding dimension. 
Let $L^{(r)} \in \mathbb{R}^{n \times n}$ denote the relation-specific semantic subspace Laplacian associated with relation type $r$. We define the projected matrix:

\begin{equation}
M_r = Z^\top L^{(r)} Z.
\end{equation}

The optimal subspace projection $P_r \in \mathbb{R}^{d\times d_r}$ is obtained by solving:

\begin{equation}
\min_{P_r^\top P_r = I_{d_r}} 
\mathrm{Tr}\left(P_r^\top Z^\top L^{(r)} Z P_r\right), 
\end{equation}
where the orthogonality constraint $P_r^\top P_r = I_{d_r}$ ensures non-redundant and mutually independent subspace directions.

According to the Courant-Fischer theorem, the solution consists of the eigenvectors corresponding to the smallest eigenvalues of $M_r$. Let $\{\lambda_j^{(r)}\}_{j=1}^{d}$ denote the ordered eigenvalues of $M_r$. The optimal subspace dimension $d_r$ is selected via the eigen-gap heuristic:

\begin{equation}
d_r = \arg\max_{1 \le j < d} 
\left(\lambda_{j+1}^{(r)} - \lambda_j^{(r)}\right).
\end{equation}

This criterion identifies the intrinsic spectral transition point, ensuring sufficient semantic expressiveness while avoiding redundant dimensions. 

To clarify the trade-off between semantic expressiveness and communication efficiency, we further emphasize the definitions of key variables involved in our subspace dimension selection. The relation-specific Laplacian associated with relation $r$ is denoted by $L^{(r)} \in \mathbb{R}^{n \times n}$, and the projection matrix for relation $r$ is defined as $P_r \in \mathbb{R}^{d\times d_r}$, subject to the orthogonality constraint $P_r^\top P_r = I_{d_r}$. The eigen-gap heuristic thus rigorously determines the optimal subspace dimension $d_r$ by identifying the largest gap in the eigenvalue spectrum of the projected Laplacian matrix $M_r=Z^\top L^{(r)}Z$. This approach provides a mathematically well-grounded criterion to balance semantic representation accuracy with practical considerations of communication overhead.

Furthermore, we emphasize that the selection of the number of semantic subspaces $K$ is governed by the cardinality of the relation set $\mathcal{R}$ rather than by the number of hyperedges. Specifically, each relation type $r\in\mathcal{R}$ naturally corresponds to a distinct semantic subspace. Hence, the total number of subspaces $K$ is determined as: $K = |\mathcal{R}|.$
This choice aligns subspace partitioning directly with semantic relation types, preserving relational coherence and semantic integrity. Moreover, this definition avoids unnecessary fragmentation of the semantic space, effectively balancing the granularity of relational semantics and computational overhead, thus ensuring efficient and robust semantic representation and inference.

The optimal semantic subspace configuration, including the number (determined by $|\mathcal{R}|$) and the dimensionality (determined by the eigen-gap heuristic), is computed offline based on static or historical semantic datasets. As a result, the intrinsic semantic geometry captured by these subspaces remains stable and does not require frequent adjustments during online deployment. Channel state variations, such as fluctuations in signal-to-noise ratio or fading conditions, predominantly influence transmission-level decisions, including quantization precision and selective prioritization of semantic subspaces, rather than altering the fundamental configuration of subspaces. Therefore, the semantic subspace construction is inherently robust and computationally efficient in dynamic channel environments.




\subsubsection{Complexity analysis}

We now can analyze the computational complexity of the proposed Semantic Subspace Construction algorithm . For each iteration of the subspace embedding procedure, the computational complexity primarily consists of two components: identifying the most semantically discrepant pairs and performing the non-linear embedding updates.

Let $|\mathcal{V}| = n$ denote the number of entities and $|\mathcal{E}| = m$ the number of hyperedges in the hypergraph. For a hyperedge $e$ with size $|e|$, identifying the most discrepant pair of entities requires computing distances for each pair of entities, which incurs an $\mathcal{O}(|e|^2 d)$ complexity, where $d$ is the embedding dimension. Summing over all hyperedges, this step yields a worst-case complexity of $\mathcal{O}(m\bar{k}^2 d)$, with $\bar{k}$ representing the average hyperedge size. Practically, this can be significantly reduced by exploiting the sparsity inherent in most real-world hypergraphs.

The second component, updating entity embeddings via the non-linear Laplacian diffusion, involves aggregating and updating embeddings for all entities, leading to a complexity of $\mathcal{O}(n d^2)$ per iteration, due to the matrix multiplications and embedding transformations involved. Hence, for a total of $L$ iterations, the overall computational complexity of the Semantic Subspace Construction is:
\begin{equation}
\mathcal{O}\left(L\left(m\bar{k}^2 d + n d^2\right)\right).
\end{equation}

Given that typically $d \ll n$ and $d \ll m\bar{k}$, the practical complexity is dominated by the hyperedge-wise pair computations and is therefore linear with respect to the number of hyperedges $m$ and quadratic with respect to average hyperedge size $\bar{k}$. Consequently, our approach remains computationally efficient and scalable for large-scale hypergraphs encountered in practical semantic communication scenarios.

\subsection{Theoretical Analysis}

We now extend the theoretical foundations of our semantic subspace construction by incorporating and adapting the formal proofs from the subspace hypergraph framework. Recall that for each entity \(v\in\mathcal{V}\) we denote its embedding by \(\mathbf{z}_v\in\mathbb{R}^{d}\), and for each incidence \((v\triangleleft e)\) we have a restriction map \(\mathbf{F}_{v\triangleleft e}\in\mathbb{R}^{d\times d}\) mapping from the vertex stalk to the hyperedge stalk. We also define \(\mathbf{D}_v=\sum_{e:v\in e}\mathbf{F}_{v\triangleleft e}^\top\mathbf{F}_{v\triangleleft e}\) as the degree-like normalization term in the stalk domain.

\subsubsection{Optimal Number of Subspaces}

Let \(Z \in \mathbb{R}^{n\times d}\) denote the semantic embedding matrix, where \(n\) is the number of entities and \(d\) the embedding dimension. We determine the optimal semantic subspace for each relation \(r\) by solving:
\begin{equation}\label{eq:subspace_opt}
\min_{P_r^\top P_r=I_{d_r}} \mathrm{Tr}\!\left(P_r^\top Z^\top L^{(r)} Z P_r\right),
\end{equation}
where \(P_r \in \mathbb{R}^{d\times d_r}\) is the projection matrix onto the corresponding semantic subspace of dimension \(d_r\).

By the Courant–Fischer theorem, the solution to \eqref{eq:subspace_opt} is given by the eigenvectors of \(M_r = Z^\top L^{(r)} Z\) associated with its smallest eigenvalues:
\begin{equation}
M_r\mathbf{v}_i=\lambda_i\mathbf{v}_i,\quad \lambda_1\leq\lambda_2\leq\dots\leq\lambda_d,\quad
P_r=[\mathbf{v}_1,\dots,\mathbf{v}_{d_r}].
\end{equation}
This guarantees minimal intra-subspace variation and thus preserves semantic coherence.

In practice, the optimal dimension \(d_r\) is chosen via the eigen-gap heuristic, selecting the point at which the decrease in eigenvalue levels off:
\begin{equation}
d_r=\arg \max_{1\leq j<d}(\lambda_{j+1}-\lambda_j).
\end{equation}
This balances expressiveness and efficiency, ensuring each subspace captures essential semantics without redundancy.

\subsubsection{subspace Dirichlet Energy} 
The subspace Dirichlet energy of a signal \(\mathbf{z}\in\mathbb{R}^{n\times d}\) on the hypergraph is defined as:

\begin{small}
\begin{equation}\label{eq:subspace_dirichlet}
E_{\mathcal{F}}^{L^2}(\mathbf{z})=\frac{1}{2}\sum_{e\in\mathcal{E}}\frac{1}{\delta_e}\sum_{u,v\in e}\bigl\|\mathbf{F}_{v\triangleleft e}\mathbf{D}_v^{-1/2}\mathbf{z}_v-\mathbf{F}_{u\triangleleft e}\mathbf{D}_u^{-1/2}\mathbf{z}_u\bigr\|_2^2,
\end{equation}
\end{small}

where \(\delta_e\) denotes the size of hyperedge \(e\). This energy measures discrepancies between neighbouring entities in the hyperedge stalk domain rather than in the raw feature space. Minimizing \eqref{eq:subspace_dirichlet} encourages consensus of representations in the more expressive stalk space, avoiding the trivial uniformity often produced by classical Laplacians.

\subsubsection{Linear subspace Laplacian Minimizes} 
Let 
\begin{equation}
(\Delta_{\mathcal{F}}= \mathbf{D}^{-1/2}\mathbf{L}_{\mathcal{F}}\mathbf{D}^{-1/2}\in\mathbb{R}^{nd\times nd}) ,
\end{equation}
denote the symmetric normalized linear subspace hypergraph Laplacian. A single diffusion step can be written as:
\begin{equation}
    \mathbf{Y}=(\mathbf{I}-\Delta_{\mathcal{F}})\mathbf{Z}.
\end{equation}

Then, we have:

\begin{equation}
E_{\mathcal{F}}^{L^2}(\mathbf{Y})\leq\lambda^*E_{\mathcal{F}}^{L^2}(\mathbf{Z}),\qquad \lambda^*=\max_i(1-\lambda_i)^2<1,
\end{equation}

where \(\{\lambda_i\}\) are the nonzero eigenvalues of \(\Delta_{\mathcal{F}}\). Iterating this process yields convergence of the subspace Dirichlet energy to zero, showing that the diffusion step acts as a contraction mapping in the stalk domain.

While the above analysis establishes the contraction property and energy decay behavior of the linear subspace Laplacian, the linear formulation still relies on pairwise aggregation over all induced connections within each hyperedge. To further improve semantic expressiveness and better preserve higher-order structural discrepancies, we next introduce the non-linear subspace formulation based on total variation.

\subsubsection{Subspace Total Variation and Nonlinear Laplacian}

For the non-linear case we define the subspace total variation as:

\begin{small}
\begin{equation}\label{eq:subspace_tv_energy}
E_{\mathcal{F}}^{TV}(\mathbf{z})=\frac{1}{2}\sum_{e\in\mathcal{E}}\frac{1}{\delta_e}\max_{u,v\in e}\bigl\|\mathbf{F}_{v\triangleleft e}\mathbf{D}_v^{-1/2}\mathbf{z}_v-\mathbf{F}_{u\triangleleft e}\mathbf{D}_u^{-1/2}\mathbf{z}_u\bigr\|_2^2,
\end{equation}
\end{small}

We can show that \(E_{\mathcal{F}}^{TV}(\mathbf{z})\) is convex but nondifferentiable and that the Laplacian operator of symmetric normalized nonlinear subspace,

\begin{footnotesize}
\begin{equation}
\Delta_{\mathcal{\bar{F}}}\mathbf{z}=\sum_{e;u\sim_ev}\frac{1}{\delta_e}\mathbf{D}_v^{-1/2}\mathbf{F}_{v\triangleleft e}^\top\bigl(\mathbf{F}_{v\triangleleft e}\mathbf{D}_v^{-1/2}\mathbf{z}_v-\mathbf{F}_{u\triangleleft e}\mathbf{D}_u^{-1/2}\mathbf{z}_u\bigr),
\end{equation}
\end{footnotesize}

acts as a subgradient of \(E_{\mathcal{F}}^{TV}(\mathbf{z})\). Therefore, the diffusion process
\begin{equation}
\mathbf{Z}^{(l+1)}=\mathbf{Z}^{(l)}-\eta\,\Delta_{\mathcal{\bar{F}}}\mathbf{Z}^{(l)},
\end{equation}
with step size \(\eta>0\) performs subgradient descent on \eqref{eq:subspace_tv_energy}, ensuring monotone decrease of the subspace total variation and convergence to a minimizer.

\subsubsection{Eigenvalue Characterization of Subspace Projections}

Given the block matrix \(M_r=\mathbf{Z}^\top L^{(r)}\mathbf{Z}\), where the optimal projection \(P_r\in\mathbb{R}^{d\times d_r}\) minimizing
\begin{equation}
    \min_{P_r^\top P_r=I}\;\mathrm{Tr}(P_r^\top \mathbf{Z}^\top L^{(r)}\mathbf{Z}P_r),
\end{equation}
is formed by the eigenvectors associated with the \(d_r\) smallest eigenvalues of \(M_r\). This ensures that each learned subspace is the one of minimum subspace Dirichlet or total variation energy, thereby preserving hyperedge cohesion while maintaining distinctiveness across different relations.

These extended proofs formally justify why our semantic subspace construction inherits the contraction properties of subspace Laplacian diffusion and why the resulting relation-specific subspaces naturally minimize subspace total variation, yielding robust and expressive higher-order semantic embeddings.

\section{Implicit Semantic Inference}\label{sec:implicit_inference}

In this section, we present the \emph{Implicit Semantic Inference} module of HISR. Building upon the semantic subspaces generated in Section~\ref{sec:semantic_subspace}, this module integrates information across relation-specific subspaces to recover missing or corrupted semantic relationships after transmission. By operating within relation-specific subspaces, the inference process preserves semantic coherence and prevents over-mixing of heterogeneous relations, which is critical for robust reasoning in noisy or incomplete settings.

\begin{figure}
    \centering
    \includegraphics[width=1\linewidth]{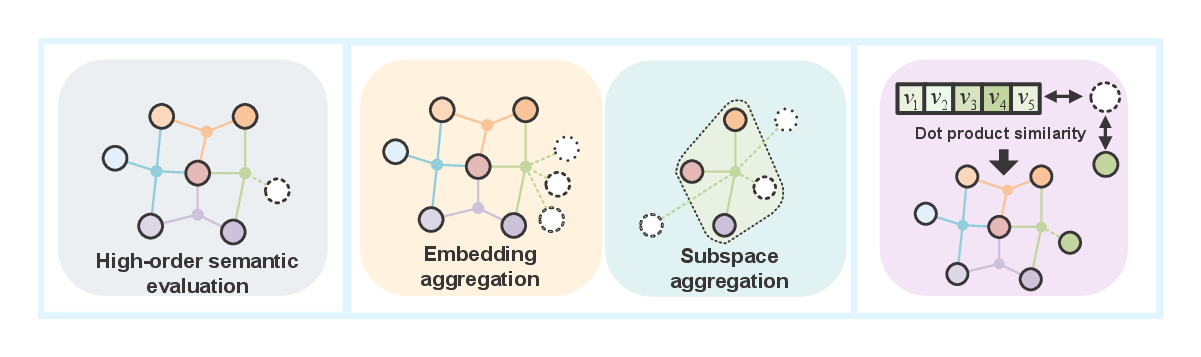}
    \caption{Example of Implicit Semantic Inference via Relation-Specific Semantic Subspaces}
    \label{inference mechanism}
\end{figure}

\subsection{Position-Aware Aggregation}

Effective semantic inference relies critically on the accurate aggregation and representation of entity embeddings within their relation-specific contexts. Traditional aggregation methods often fail to capture positional semantic distinctions, resulting in ambiguous or overly uniform embeddings. To overcome this limitation, we propose a position-aware aggregation mechanism designed explicitly to leverage the unique semantic contributions of each entity according to its relational position within hyperedges.  

Formally, let \(\mathcal{S}_r\) denote the semantic subspace associated specifically with relation type \(r\). Consider a $k$-ary hyperedge (semantic tuple) denoted as \(\tau = r(e_1, e_2, \dots, e_k)\), where each entity \(e_p\) occupies a distinct positional role. Within the relation-specific semantic subspace \(\mathcal{S}_r\), we define the positional-aware semantic aggregation for an entity \(e_i\) as follows:
\begin{equation}\label{eq:position_aware_agg}
\mathbf{n}_{e_i \leftarrow \tau}^{(r)} = \sum_{\substack{p=1 \\ p\neq i}}^{k}\mathbf{z}_{e_p}^{(r)}\circ\mathbf{r}_p^{(r)},
\end{equation}
where the embedding vector \(\mathbf{z}_{e_p}^{(r)}\in\mathbb{R}^{d}\) represents the semantic embedding of entity \(e_p\) within the dedicated subspace \(\mathcal{S}_r\). The vector \(\mathbf{r}_p^{(r)}\in\mathbb{R}^{d}\) explicitly encodes position-specific semantic characteristics corresponding to the particular role of entity \(e_p\) in the relational context, thus differentiating the semantic contributions of each position within the tuple. Here, \(\circ\) denotes the Hadamard (element-wise) product operation, ensuring precise and efficient combination of positional semantics.

The rationale behind the position-specific vector \(\mathbf{r}_p^{(r)}\) is two-fold. First, it enhances semantic specificity by highlighting role-dependent entity contributions within each hyperedge. For instance, in a tuple expressing an event, the positional vectors allow entities in different roles (such as agent, object, and recipient) to contribute uniquely and distinctly to the overall semantic representation. Second, this positional encoding explicitly prevents semantic ambiguity by ensuring that entities' semantic embeddings are precisely modulated according to their relational positions, rather than being uniformly aggregated. This mechanism effectively retains rich semantic distinctions and prevents embedding collapse or over-smoothing, common in conventional semantic aggregation schemes.

\subsection{Subspace-Guided Aggregation and Iterative Refinement}

Once relation-specific semantic subspaces have been constructed, the next step is to refine and update entity representations in a manner that fully exploits the structured information captured by these subspaces. This process involves two tightly coupled operations: (i) aggregating position-aware contextual information from all hyperedges in which an entity participates, and (ii) iteratively refining and updating the entity embeddings to achieve coherent, relation-specific representations. Together, these operations ensure that semantic reasoning is performed entirely within the correct subspace, thereby preventing the mixing of unrelated relational contexts and enhancing both accuracy and robustness.

For each entity $e_i$ and each relation-specific semantic subspace $\mathcal{S}_r$, the local context is aggregated across all tuples involving $e_i$:
\begin{equation}\label{eq:refine_context}
\mathbf{n}_{e_i}^{(r)}=\sigma\Bigg(\sum_{\tau\in\mathcal{N}_{e_i}}W_1\,\mathbf{n}_{e_i\leftarrow\tau}^{(r)}\Bigg),
\end{equation}
where $\mathcal{N}_{e_i}$ denotes the set of tuples containing $e_i$, $W_1\in\mathbb{R}^{d\times d}$ is a learned transformation matrix, and $\sigma(\cdot)$ is a non-linear activation such as ReLU. This formulation ensures that information aggregated for $e_i$ originates strictly from the same subspace $\mathcal{S}_r$, thereby preserving semantic purity.

To further integrate self-identity and prevent semantic drift during aggregation, we introduce a virtual relation embedding $\mathbf{r}_v\in\mathbb{R}^{d}$, which is combined with the refined context:
\begin{equation}\label{eq:refined_embedding}
\hat{\mathbf{z}}_{e_i}^{(r)}=\sigma\Bigg(
W_1\sum_{\tau\in\mathcal{N}_{e_i}}\sum_{p=1}^{k}\mathbf{z}_{e_p}^{(r)}\circ\mathbf{r}_p^{(r)}+W_2\bigl(\mathbf{z}_{e_i}^{(r)}\circ\mathbf{r}_v\bigr)\Bigg),
\end{equation}
where $W_2\in\mathbb{R}^{d\times d}$ is an additional learnable transformation, and $\circ$ denotes the Hadamard product. The explicit inclusion of $\mathcal{S}_r$ in \eqref{eq:refined_embedding} guarantees that updates are performed only within the correct relation-specific subspace, preventing cross-subspace contamination.

Rather than performing a single-pass update, HISR refines entity embeddings iteratively so that the representations progressively converge to a stable and semantically coherent point within each subspace. At iteration $t+1$, the update rule for the embedding of $e_i$ in subspace $\mathcal{S}_r$ is:

\begin{small}

\begin{align}\label{eq:iterative_update}
\hat{\mathbf{z}}_{e_i}^{(r),(t+1)}=\sigma\Bigg(
& W_1^{(t)}\sum_{\tau\in\mathcal{N}_{e_i}}\sum_{\substack{p=1\\p\neq i}}^{k}\hat{\mathbf{z}}_{e_p}^{(r),(t)}\circ\mathbf{r}_p^{(r),(t)} \nonumber \\
& + W_2^{(t)}\bigl(\hat{\mathbf{z}}_{e_i}^{(r),(t)}\circ\mathbf{r}_v^{(t)}\bigr)\Bigg),
\end{align}

\end{small}

where $W_1^{(t)}$ and $W_2^{(t)}$ are layer-specific transformations. By iterating \eqref{eq:iterative_update}, entity embeddings gradually align with the semantic structure encoded by their relation-specific subspaces, thereby improving intra-subspace coherence and inter-subspace separation.

This unified subspace-guided aggregation and iterative refinement strategy ensures that each entity representation not only incorporates all relevant context but also remains semantically faithful to its specific relational role. As a result, HISR achieves more robust, high-fidelity semantic reasoning and recovery even under adverse transmission conditions or incomplete information scenarios.

To handle scenarios with incomplete or corrupted relational information during semantic subspace construction, we introduce a relation-level dropout mechanism to explicitly model partial knowledge conditions. Let $H = (\mathcal{V}, \mathcal{E}, \mathcal{R})$ denote the semantic hypergraph. During training, we define a random masking variable $\xi_e \sim \text{Bernoulli}(1-\rho)$ for each hyperedge $e \in \mathcal{E}$, where $\rho \in [0,1)$ denotes the dropout probability. The effective hyperedge set used for subspace construction is therefore
\begin{equation}
\mathcal{E}^{(\rho)} = \{ e \in \mathcal{E} \mid \xi_e = 1 \},
\end{equation}

The subspace Laplacian $L^{(r)}$ is then constructed over the reduced hypergraph $H^{(\rho)} = (\mathcal{V}, \mathcal{E}^{(\rho)}, \mathcal{R})$, and the corresponding projected matrix becomes
\begin{equation}
M_r^{(\rho)} = Z^\top L^{(r,\rho)} Z,
\end{equation}
where $L^{(r,\rho)}$ denotes the relation-specific semantic subspace Laplacian associated with relation type $r$ and computed on the masked hypergraph $H^{(\rho)}$.

By optimizing the subspace projections with respect to randomly perturbed relational structures, the model implicitly minimizes the expected subspace energy
\begin{equation}
\mathbb{E}_{\xi}\!\left[\mathrm{Tr}\!\left(P_r^\top Z^\top L^{(r,\rho)} Z P_r\right)\right],
\end{equation}
which encourages semantic embeddings that are stable under structural perturbations. This stochastic regularization prevents the learned subspaces from overfitting to fully observed relational patterns and improves generalization when relational information is missing, corrupted, or partially decoded during physical channel transmission. Consequently, the learned semantic subspace mappings remain robust and discriminative even under incomplete or noisy relational contexts.

\subsection{Subspace-Guided Inference Rule}

Given the refined embeddings, we define a position-aware neural scoring function to evaluate the credibility of a candidate $k$-tuple:
\begin{align}\label{eq:pos_score}
f_r^{(p)}(e_1,\dots,e_k)=
&\sigma\Big(\mathrm{vec}\Big(\sum_{\substack{j=1\\j\neq p}}^{k}\Theta_r(\hat{\mathbf{z}}_{e_j}^{(r)}\circ\mathbf{r}_j^{(r)},j)\Big)Y\Big)\nonumber \\
&\cdot\Theta_r(\hat{\mathbf{z}}_{e_p}^{(r)}\circ\mathbf{r}_p^{(r)},p),
\end{align}
where $\Theta_r(\cdot,p)$ denotes a position-dependent shift operator, $Y$ is a learned linear transformation, and $\mathrm{vec}(\cdot)$ vectorizes its input.  

The final score of the candidate hyperedge is obtained by averaging over positions:
\begin{equation}\label{eq:final_score}
f_r(e_1,\dots,e_k)=\frac{1}{k}\sum_{p=1}^{k}f_r^{(p)}(e_1,\dots,e_k).
\end{equation}

Finally, the reconstructed hyperedge set $\hat{\mathcal{E}}$ at the receiver is:
\begin{equation}\label{eq:reconstruct}
\hat{\mathcal{E}}=\Bigl\{e: \sum_{r\in\mathcal{R}}\beta_r f_r(e)\geq\gamma\Bigr\},
\end{equation}
where $\beta_r$ are normalization weights for different relations and $\gamma$ is a global threshold controlling the acceptance of inferred hyperedges.

This formulation shows explicitly how relation-specific subspaces guide the entire inference pipeline. All aggregation, refinement, scoring, and reconstruction operations are performed within their respective subspaces, ensuring that higher-order semantics are faithfully captured and recovered even under noise or partial transmission loss.

To robustly reconstruct implicit semantic information at the receiver, we design an inference mechanism that explicitly leverages relation-specific semantic subspaces. Given a received partial tuple \(\mathcal{T}\), our goal is to identify and complete missing entities by exploring their potential semantic coherence within corresponding subspaces. 

Formally, let \(\mathcal{L}_{\tau}\) denote the set of candidate latent entities associated with an incomplete hyperedge (tuple) \(\mathcal{T}\). For each candidate latent entity \( e \in \mathcal{L}_{\tau} \), we first project its embedding into the relation-specific semantic subspace corresponding to \(\mathcal{T}\). The semantic credibility score \(C(e, \mathcal{T})\) of entity \(e\) with respect to tuple \(\mathcal{T}\) is computed via the previously defined semantic scoring function \eqref{eq:final_score} as follows:
\begin{equation}\label{eq:credibility}
C(e, \mathcal{T}) = \sum_{r\in\mathcal{R}} \beta_r f_r(e, \mathcal{T}),
\end{equation}
where \(\beta_r\) represents normalization weights, and \(f_r(e,\mathcal{T})\) is computed based on embeddings projected in the subspace associated with relation \(r\).

Once semantic credibility scores are obtained, the inference algorithm identifies the most probable latent entity \(\mathcal{L}^*\) according to:
\begin{equation}\label{eq:max_credibility}
\mathcal{L}^* = \arg\max_{e \in \mathcal{L}_\tau} C(e, \mathcal{T}),
\end{equation}
and subsequently utilizes this entity for semantic completion of the incomplete tuple.

Fig.~\ref{inference mechanism} provides an illustrative example of our proposed semantic inference mechanism based on relation-specific semantic subspaces. Initially (top panel), a query is formulated, aiming to identify a missing entity (denoted by the dashed entity) within a particular semantic hyperedge $e_1$, associated with entities such as Neil Armstrong and Buzz Aldrin. Candidate entities (Michael Collins, Jim Lovell, and John Glenn) are identified for completion. 

The middle panel demonstrates the critical step of subspace mapping. Entities and their relationships are projected into the correct semantic subspace corresponding to the hyperedge $e_1$. Within this dedicated subspace, candidate entities undergo semantic evaluation. The correct subspace effectively highlights Michael Collins as a highly compatible entity, resulting in accurate reasoning. Conversely, when entities are projected into an incorrect semantic subspace (right sub-panel), meaningful reasoning fails due to semantic mismatch, illustrating the importance of accurate subspace selection.

Finally, the bottom panel depicts credibility scoring, in which semantic compatibility of each candidate entity is quantitatively evaluated. Michael Collins achieves the highest credibility score, confirming his correct identification as the missing entity.

The complete inference procedure is summarized in Algorithm~\ref{alg:semantic_inference}:

\begin{algorithm}[ht]
\LinesNotNumbered
\caption{HISR Inference Mechanism}
\label{alg:semantic_inference}

\KwIn{Hypergraph Knowledge Base $\mathcal{KB}=(\mathcal{E},\mathcal{R},\mathcal{T})$, received partial tuple $\mathcal{T}$, semantic subspaces $\{\mathcal{S}_r\}$}
\KwOut{Completed semantic tuple $\mathcal{T}'$}

Initialize candidate latent entity set $\mathcal{L}_{\tau}$ associated with tuple $\mathcal{T}$\;

\For{each candidate entity $e\in\mathcal{L}_{\tau}$}{
    Identify semantic subspace $\mathcal{S}_r$ corresponding to tuple relation type $r$\;
    Compute embedding $\mathbf{z}_{e}^{(r)}$ for candidate entity $e$ in subspace $\mathcal{S}_r$\;
    Compute semantic credibility score $C(e,\mathcal{T})$ using Eq.~\eqref{eq:credibility}\;
}

Determine most credible entity $\mathcal{L}^*$ using Eq.~\eqref{eq:max_credibility}\;
Complete the tuple: $\mathcal{T}' \gets \mathcal{T} \cup \{\mathcal{L}^*\}$\;
\Return{Completed semantic tuple $\mathcal{T}'$\;}

\end{algorithm}

This inference approach explicitly exploits subspace embeddings to accurately complete and reconstruct semantic tuples, greatly enhancing robustness to transmission errors and information loss.

\subsection{Complexity and Communication Cost Analysis}

We now analyze the computational complexity and communication overhead associated with HISR. The computational complexity of semantic inference primarily involves evaluating credibility scores and identifying optimal completions. For each incomplete tuple with candidate latent entities, the complexity of projecting embeddings into the relevant subspace and calculating the credibility score is $\mathcal{O}(|\mathcal{L}_{\tau}| d^2)$ per tuple, where $|\mathcal{L}_{\tau}|$ is the number of candidate entities and $d$ the embedding dimension. Given $|\mathcal{T}|$ tuples requiring inference, the overall inference complexity is:
\begin{equation}
\mathcal{O}\left(|\mathcal{T}||\mathcal{L}_{\tau}| d^2\right).
\end{equation}

For semantic subspace construction, complexity arises from embedding updates and hyperedge aggregation. With $n$ entities, $m$ hyperedges, average hyperedge size $\bar{k}$, and embedding dimension $d$, the complexity per iteration is dominated by aggregation operations and computed as:
\begin{equation}
\mathcal{O}(m\bar{k}^2 d + n d^2).
\end{equation}

Regarding communication cost, the number of bits required for transmitting semantic embeddings is determined by the number of subspaces and embedding dimension. The communication cost per message transmission is expressed as:
\begin{equation}\label{eq:comm_cost}
B_{\text{HISR}}=\sum_{r\in\mathcal{R}}m_r(d_r q+h_r).
\end{equation}
where $m_r$ denotes the number of embeddings transmitted per subspace associated with relation $r$, $q$ is the quantization bit-width used for embedding representation, and $h_r$ is an additional small overhead per subspace for indexing purposes.

To  characterize the trade-off between semantic expressiveness and communication efficiency, the subspace dimension selection is performed under a communication constraint. Specifically, we consider the following constrained formulation:

\begin{equation}
\min_{\{P_r\}}
\sum_{r\in\mathcal{R}}
\mathrm{Tr}\!\left(P_r^\top Z^\top L^{(r)} Z P_r\right)
\ \text{s.t.}\
B_{\text{HISR}}\le B_{\max}
\end{equation}

where $B_{\text{HISR}} = \sum_{r\in\mathcal{R}} m_r (d_r q + h_r)$ denotes the transmission overhead defined in (39), and $B_{\max}$ is the maximum allowable communication budget. This constrained optimization ensures that the selected subspace dimensions balance semantic reconstruction accuracy and practical transmission cost.

In practice, as shown empirically, increasing the number of semantic subspaces initially improves semantic accuracy but subsequently leads to diminishing returns and increased communication overhead. Hence, the optimal number of subspaces is typically selected through empirical evaluation using criteria such as the eigen-gap heuristic, balancing expressiveness and transmission efficiency.

In summary, the proposed semantic inference mechanism within relation-specific subspaces provides robust semantic completion while maintaining manageable computational complexity and communication overhead, demonstrating significant practical utility for real-world semantic communication applications.

\section{Experimental Results}

In this section, we evaluate the performance of the proposed HISR framework through extensive experiments. Specifically, we investigate the semantic inference accuracy, robustness under various channel conditions, sensitivity with respect to the number of semantic subspaces, and the impact of embedding dimension on model performance. Experiments are conducted on two representative datasets—Cora and Pubmed—under diverse scenarios, including additive Gaussian and Rayleigh noise conditions, varied subspace counts, and different embedding dimensions. The primary goal of these experiments is to rigorously demonstrate the effectiveness, robustness, and practical applicability of our semantic subspace construction and implicit inference mechanisms within realistic semantic communication contexts. To align experiments with realistic semantic communication scenarios, the transmitter only conveys compact semantic embeddings derived from local observations rather than complete hypergraphs. The receiver leverages a pre-shared knowledge base to reconstruct implicit semantics. Our benchmark datasets effectively simulate practical conditions, enabling rigorous evaluation of the robustness and accuracy of the proposed framework.

\begin{figure}
  \begin{minipage}{0.48\linewidth}
   \centering
   \includegraphics[width=4.5cm]{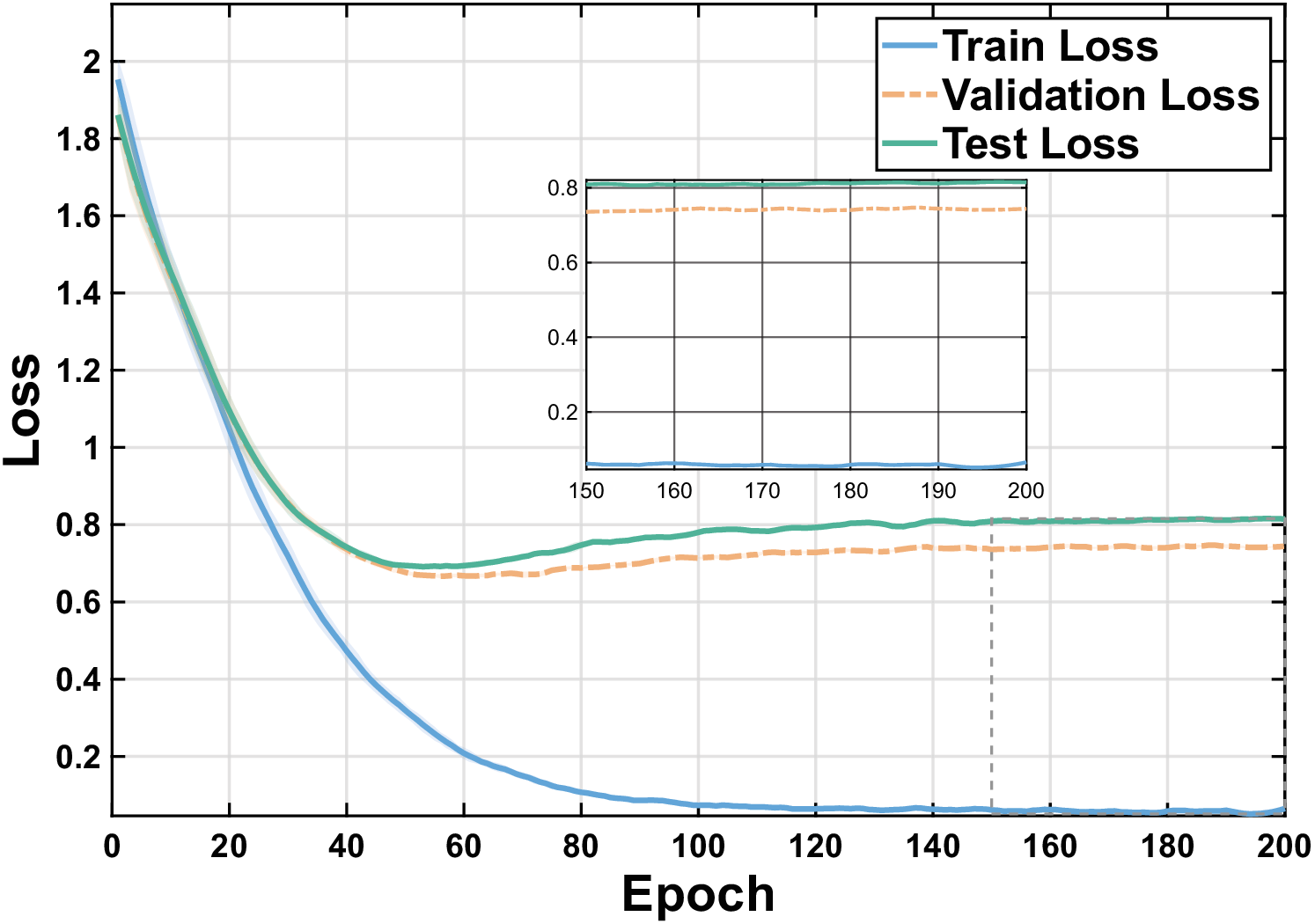}
  \caption{\small{Training, validation, and test loss curves of HISR on the Cora dataset.}}
  \label{loss-cora}
  \end{minipage}
  \hfill
  \begin{minipage}{0.48\linewidth}
   \centering
   \includegraphics[width=4.5cm]{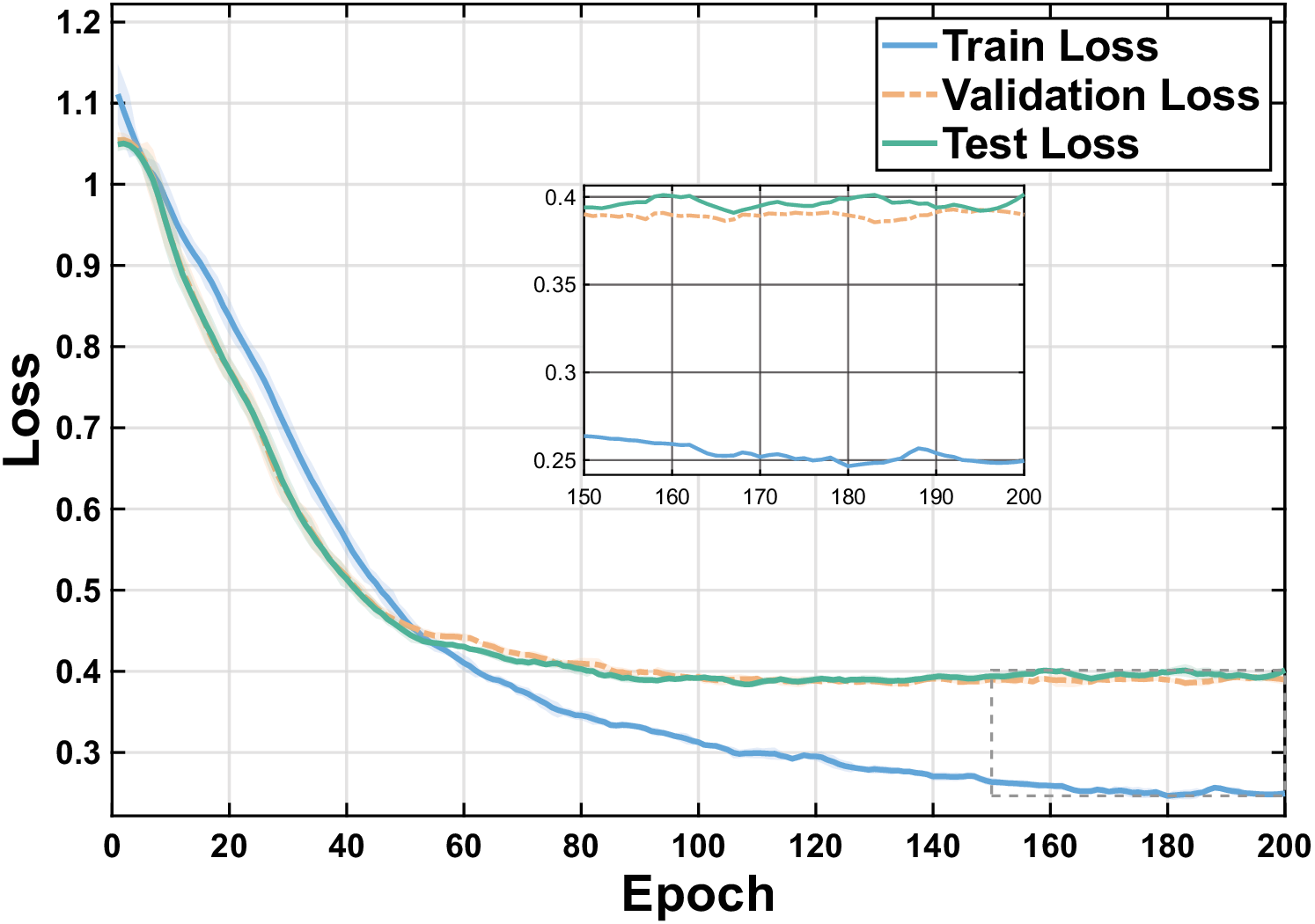}
  \caption{\small{Training, validation, and test loss curves of HISR on the Pubmed dataset.}}
  \label{loss-pubemd}
  \end{minipage}
\end{figure}

Fig.~\ref{loss-cora} and Fig.~\ref{loss-pubemd} illustrate the convergence behavior of the proposed HISR method on the Cora and Pubmed datasets, respectively. The curves depict the evolution of training, validation, and test losses across 200 epochs. On both datasets, the training loss (solid blue line) consistently decreases and converges smoothly, demonstrating stable and efficient optimization of the semantic model. Validation (dashed orange line) and test losses (solid green line) similarly exhibit steady convergence and ultimately stabilize at low values, indicating effective generalization and reliable semantic inference capabilities of HISR. The convergence patterns shown in these figures clearly validate that our method achieves robust and efficient training dynamics without noticeable overfitting, underscoring the practical effectiveness of our semantic subspace construction and inference mechanisms.

\begin{figure}
  \begin{minipage}{0.48\linewidth}
   \centering
   \includegraphics[width=4.5cm]{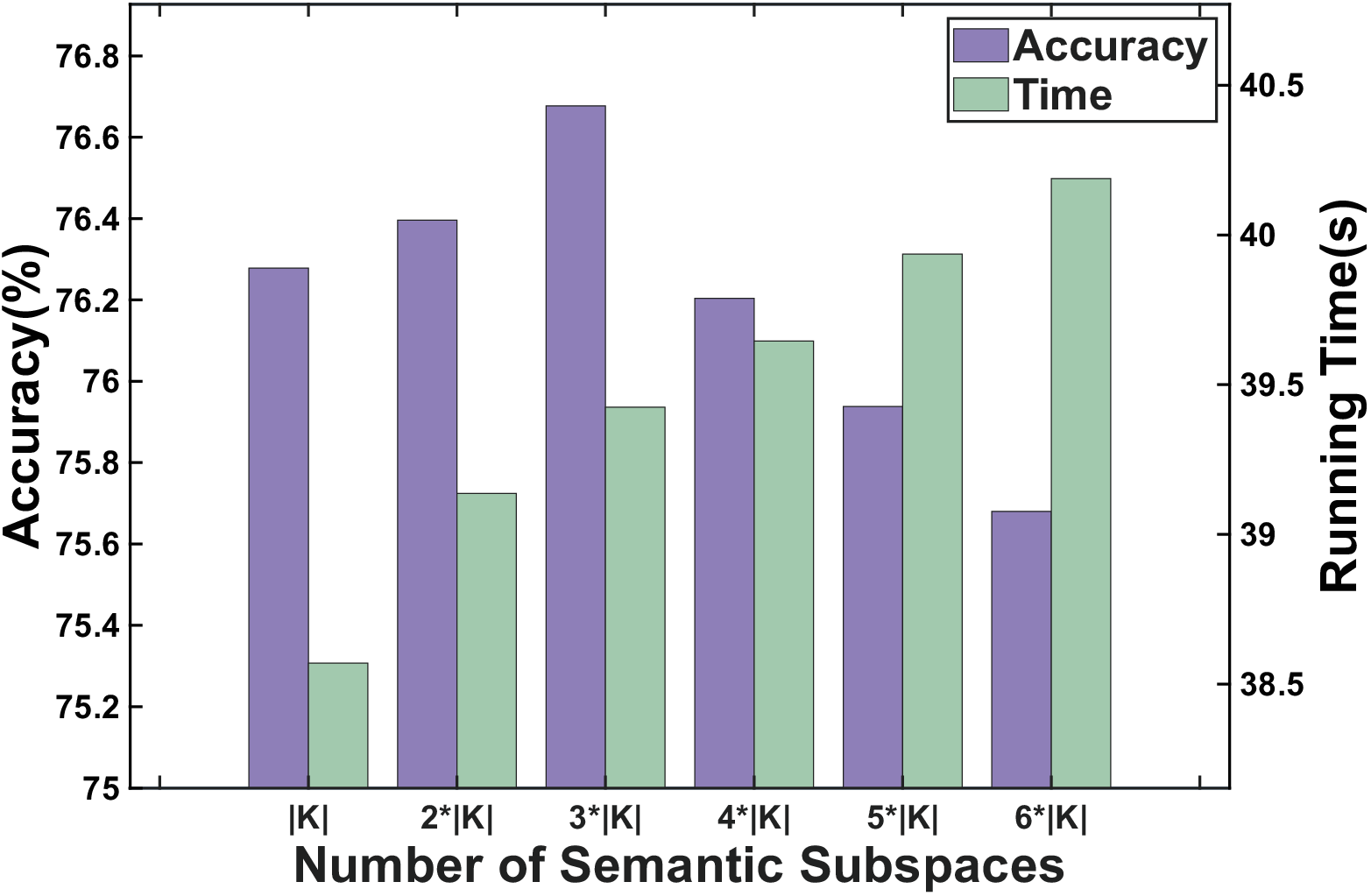}
  \caption{\small{Reasoning accuracy and running time under different numbers of semantic subspaces on the Cora dataset.}}
  \label{subspace-cora}
  \end{minipage}
  \hfill
  \begin{minipage}{0.48\linewidth}
   \centering
   \includegraphics[width=4.5cm]{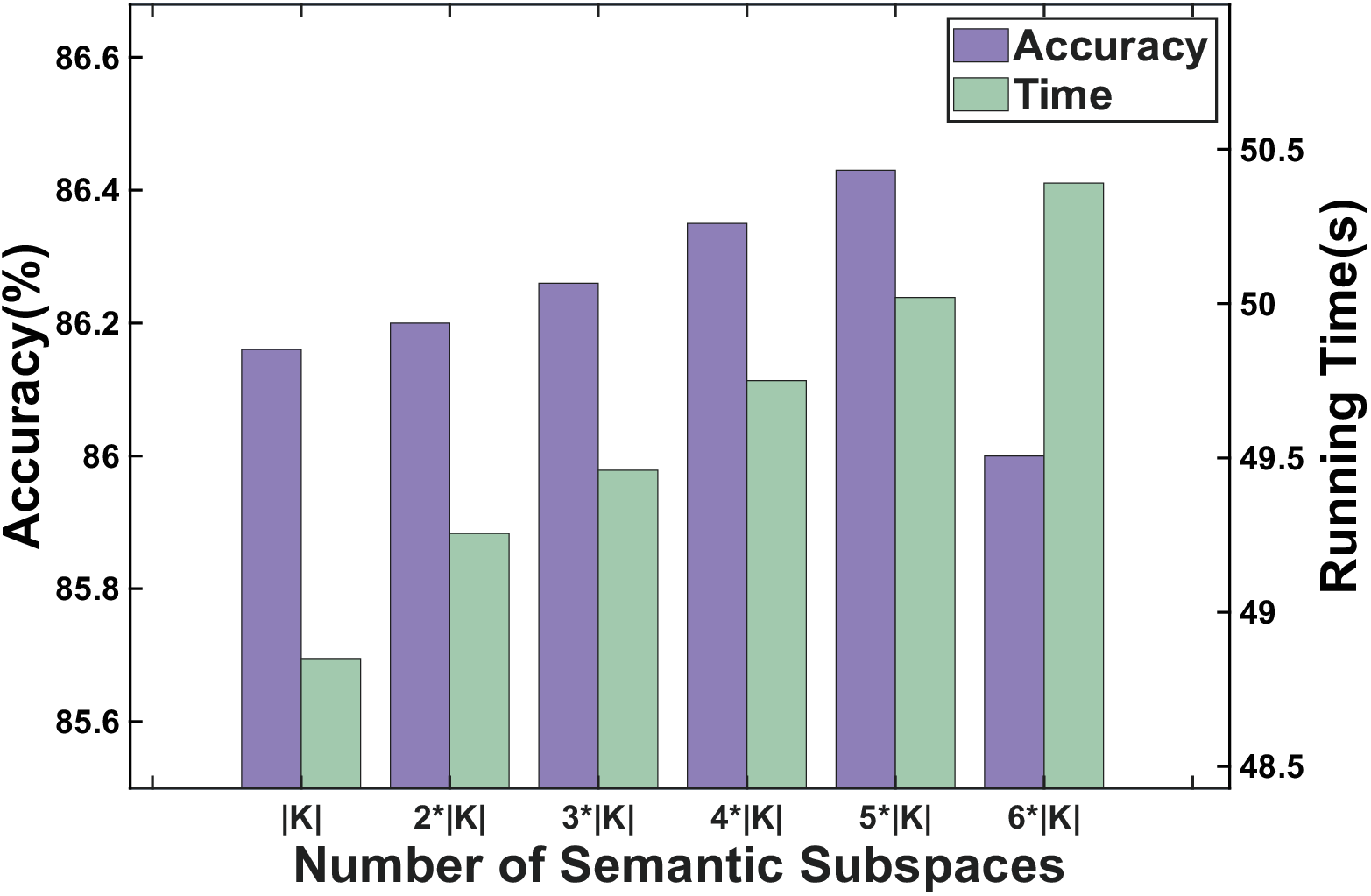}
  \caption{\small{Reasoning accuracy and running time under different numbers of semantic subspaces on the Pubmed dataset.}}
  \label{subspace-pubmed}
  \end{minipage}
\end{figure}

Fig.~\ref{subspace-cora} and Fig.~\ref{subspace-pubmed} depict the effect of varying the number of semantic subspaces on reasoning accuracy and running time for the Cora and Pubmed datasets, respectively. In both figures, the purple bars represent reasoning accuracy, while the green bars correspond to the average running time per inference.

For the Cora dataset (Fig.~\ref{subspace-cora}), reasoning accuracy initially improves as the number of semantic subspaces increases from $K$ to $K$, reaching the highest accuracy at $K$ subspaces. Beyond this point, additional subspaces provide diminishing gains and may slightly reduce accuracy, while the running time steadily increases due to the larger number of subspace projections and computations.

A similar trend is observed on the Pubmed dataset (Fig.~\ref{subspace-pubmed}). Accuracy improves with more semantic subspaces and peaks around $K$–$4K$, after which it remains stable or slightly decreases. Meanwhile, the running time exhibits a gradual upward trend with the increase of subspace count. These results confirm that the proposed relation-specific subspace mechanism effectively enhances reasoning accuracy up to an optimal point, after which the added computational overhead outweighs the benefits, highlighting the importance of selecting an appropriate number of subspaces for balancing performance and efficiency.

\begin{figure}
  \begin{minipage}{0.48\linewidth}
   \centering
   \includegraphics[width=4.5cm]{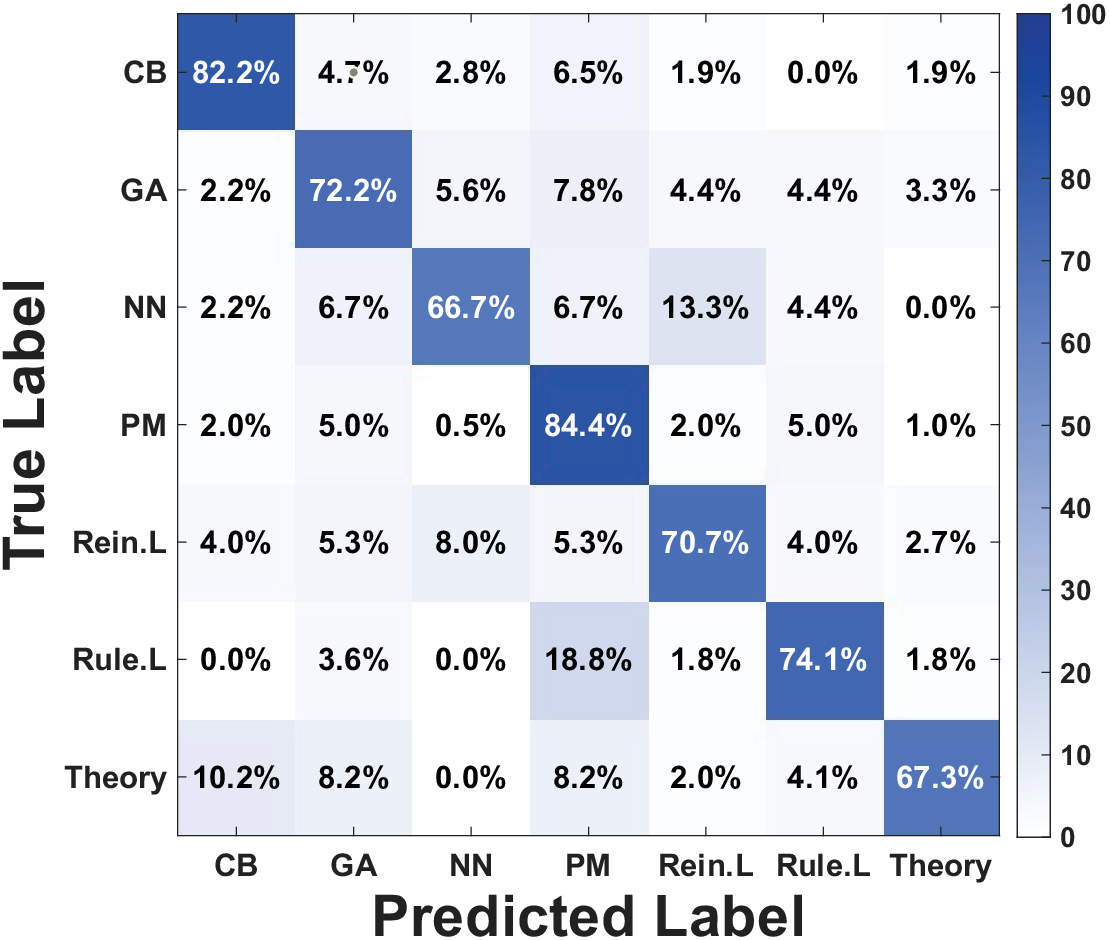}
  \caption{\small{Performance analysis on the Cora dataset.}}
  \label{Matrix-cora}
  \end{minipage}
  \hfill
  \begin{minipage}{0.48\linewidth}
   \centering
   \includegraphics[width=4.5cm]{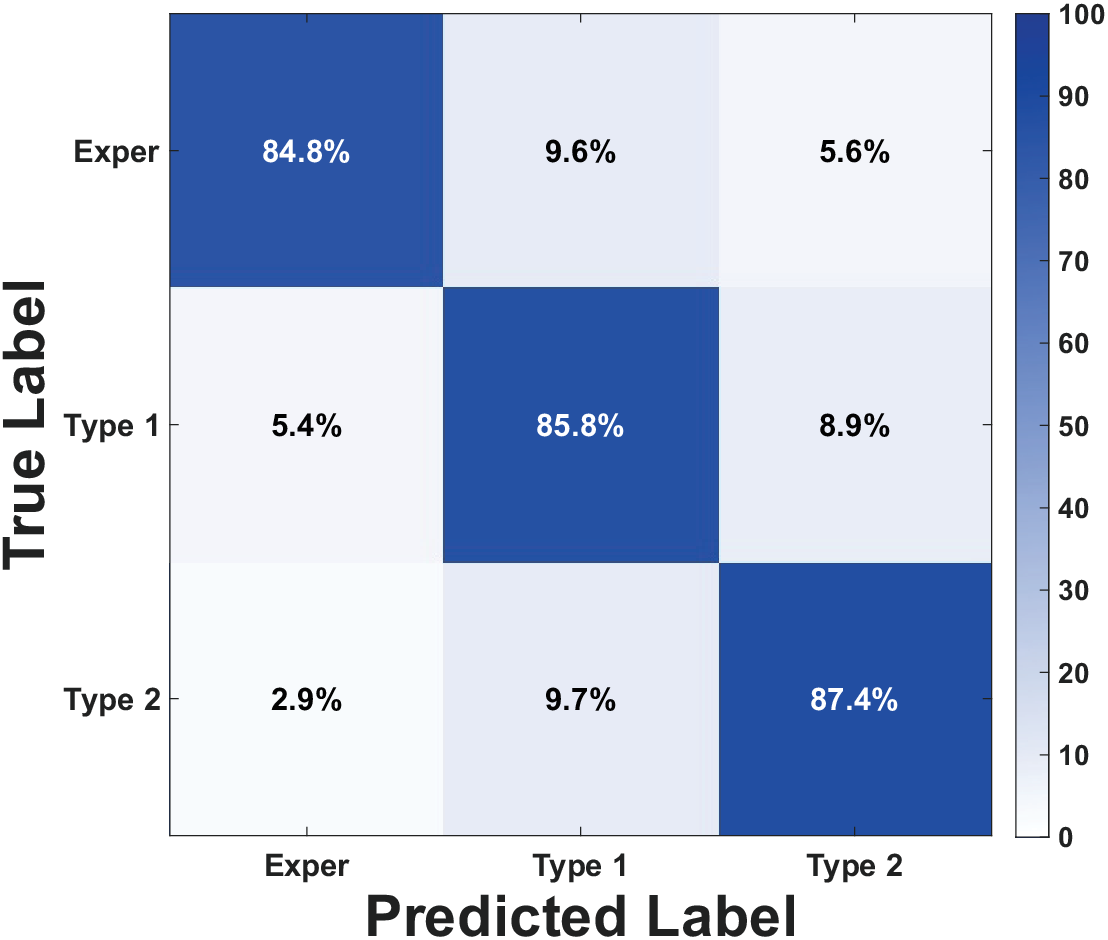}
  \caption{\small{Performance analysis on the Pubmed dataset.}}
  \label{Matrix-pubemd}
  \end{minipage}
\end{figure}

Fig.~\ref{Matrix-cora} and Fig.~\ref{Matrix-pubemd} present the confusion matrices illustrating the classification performance of HISR on the Cora and Pubmed datasets, respectively. Each cell in these matrices indicates the percentage of instances where true labels (vertical axis) are correctly or incorrectly predicted as the labels shown on the horizontal axis.

In Fig.~\ref{Matrix-cora} (Cora dataset), we observe high diagonal values, particularly notable for the categories PM (84.4\%), NN (66.7\%), and Rule.L (74.1\%), indicating strong classification accuracy in these categories. Although some misclassifications occur, they are generally minor and concentrated around semantically similar categories, reflecting the model's effectiveness in capturing complex semantic relationships.

Fig.~\ref{Matrix-pubemd} (Pubmed dataset) similarly demonstrates strong predictive performance, with diagonal accuracies consistently above 84\%, achieving a peak accuracy of 87.4\% for Type 2. Misclassifications are infrequent and minor, highlighting that the proposed HISR framework robustly distinguishes between distinct semantic categories, further underscoring the efficacy of semantic subspace embedding and inference mechanisms in achieving reliable semantic predictions.

\begin{figure}
  \begin{minipage}{0.48\linewidth}
   \centering
   \includegraphics[width=4.2cm]{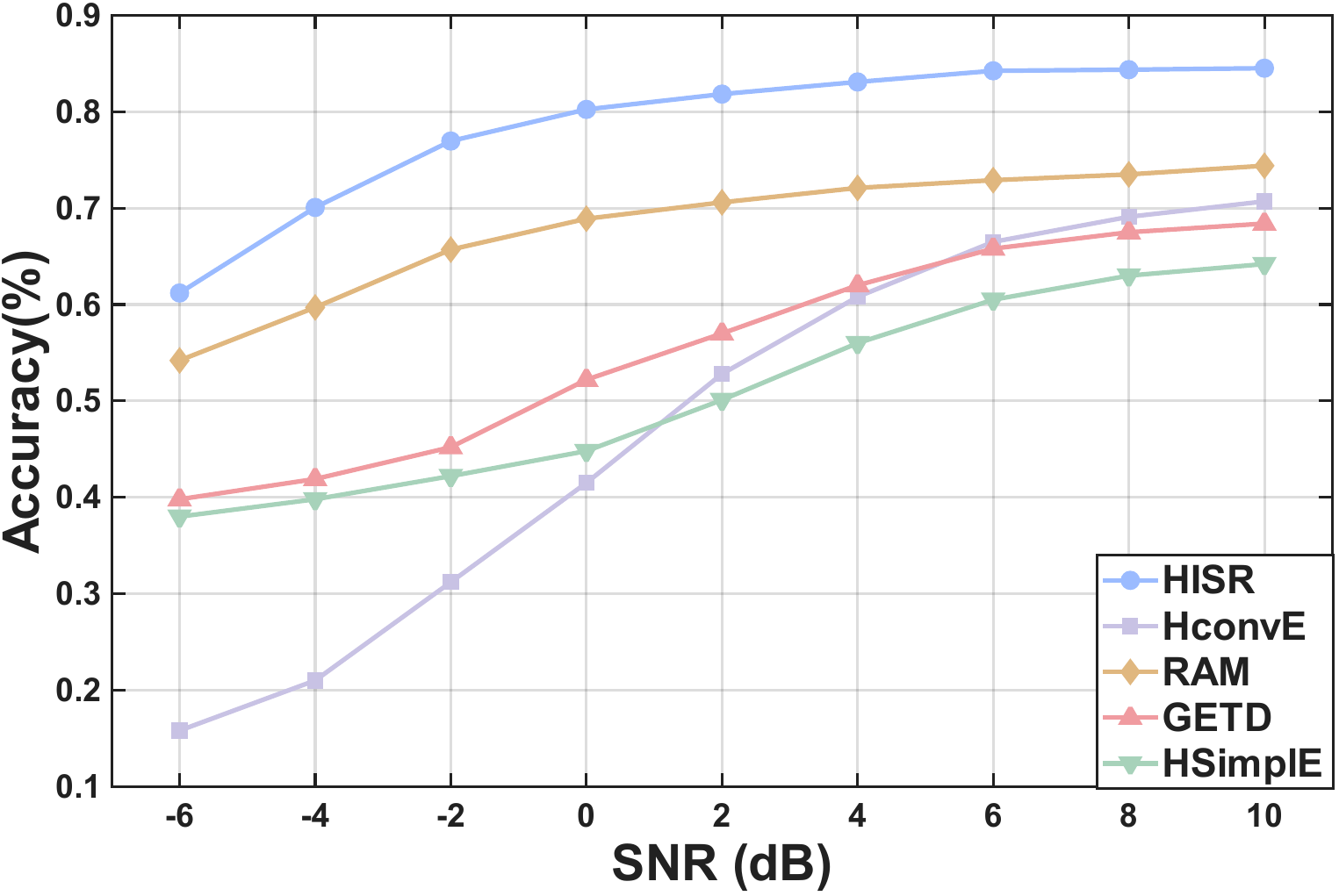}
  \caption{\small{Inference accuracy of HISR under different SNR on the FB-Auto dataset.}}
  \label{Inference-Acc-FBauto}
  \end{minipage}
  \hfill
  \begin{minipage}{0.48\linewidth}
   \centering
   \includegraphics[width=4.2cm]{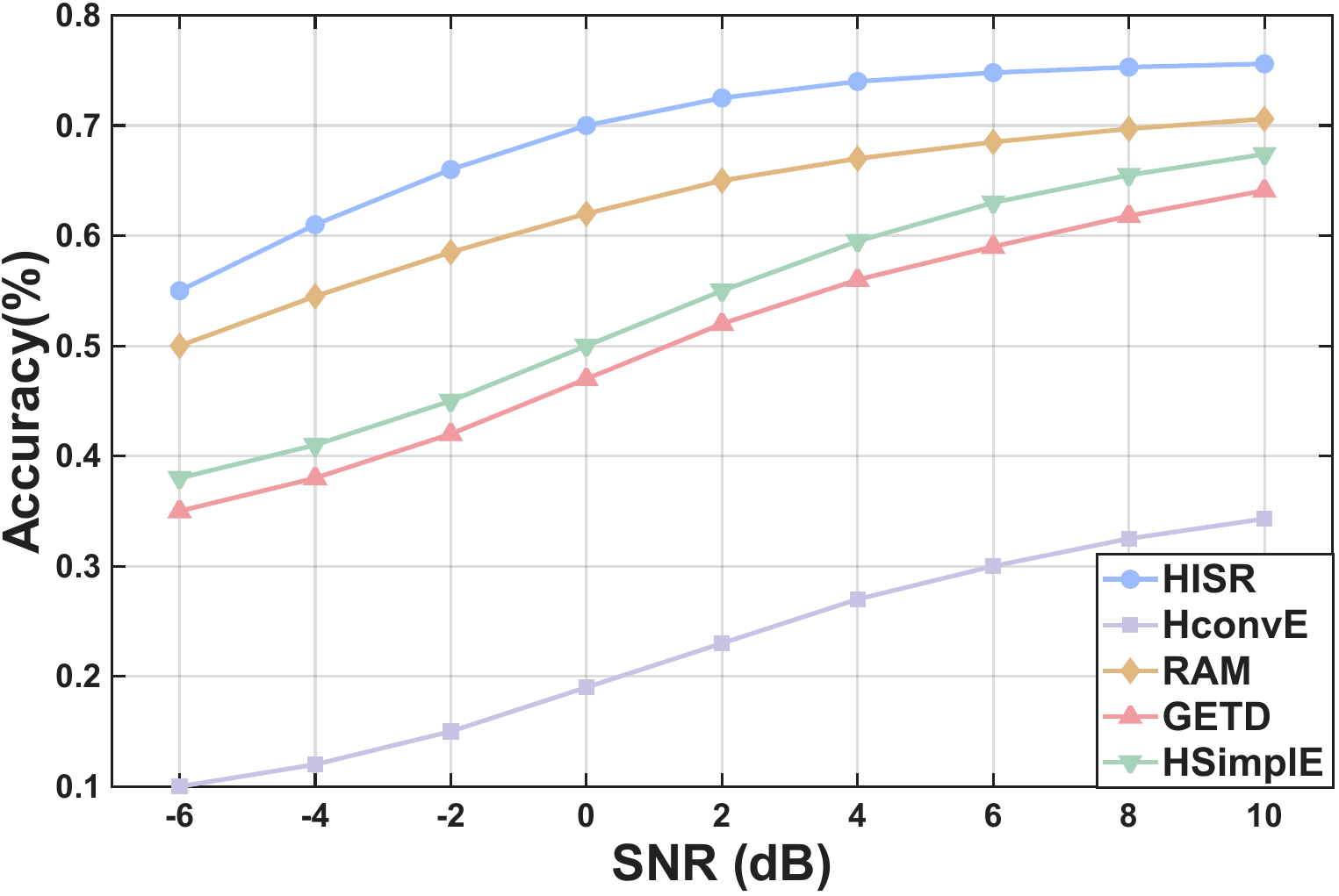}
  \caption{\small{Inference accuracy of HISR under different SNR on the M-FB15K dataset.}}
  \label{Inference-Acc-M-FB15K}
  \end{minipage}
\end{figure}

Fig.~\ref{Inference-Acc-FBauto} and Fig.~\ref{Inference-Acc-M-FB15K} illustrate the semantic inference accuracy of our proposed HISR method compared with other state-of-the-art semantic embedding models (HconvE, RAM, GETD, and HSimplE) under varying signal-to-noise ratio (SNR) conditions on the FB-Auto and M-FB15K datasets. In both figures, inference accuracy consistently increases with higher SNR, highlighting the sensitivity of semantic recovery performance to channel noise. Notably, HISR significantly outperforms all other comparative methods across the entire SNR range, maintaining an accuracy lead of approximately 7\% to 13\%, achieving superior accuracy even under extremely low SNR conditions(e.g., at -6dB, HISR achieves 61.2\% accuracy, while the best baseline only reaches 54.8\%). This demonstrates the robustness and noise resilience of the HISR framework, further underscoring its effectiveness in reliably capturing and reconstructing complex semantic relationships under challenging transmission environments.

\begin{figure}
  \begin{minipage}{0.48\linewidth}
   \centering
   \includegraphics[width=4.2cm]{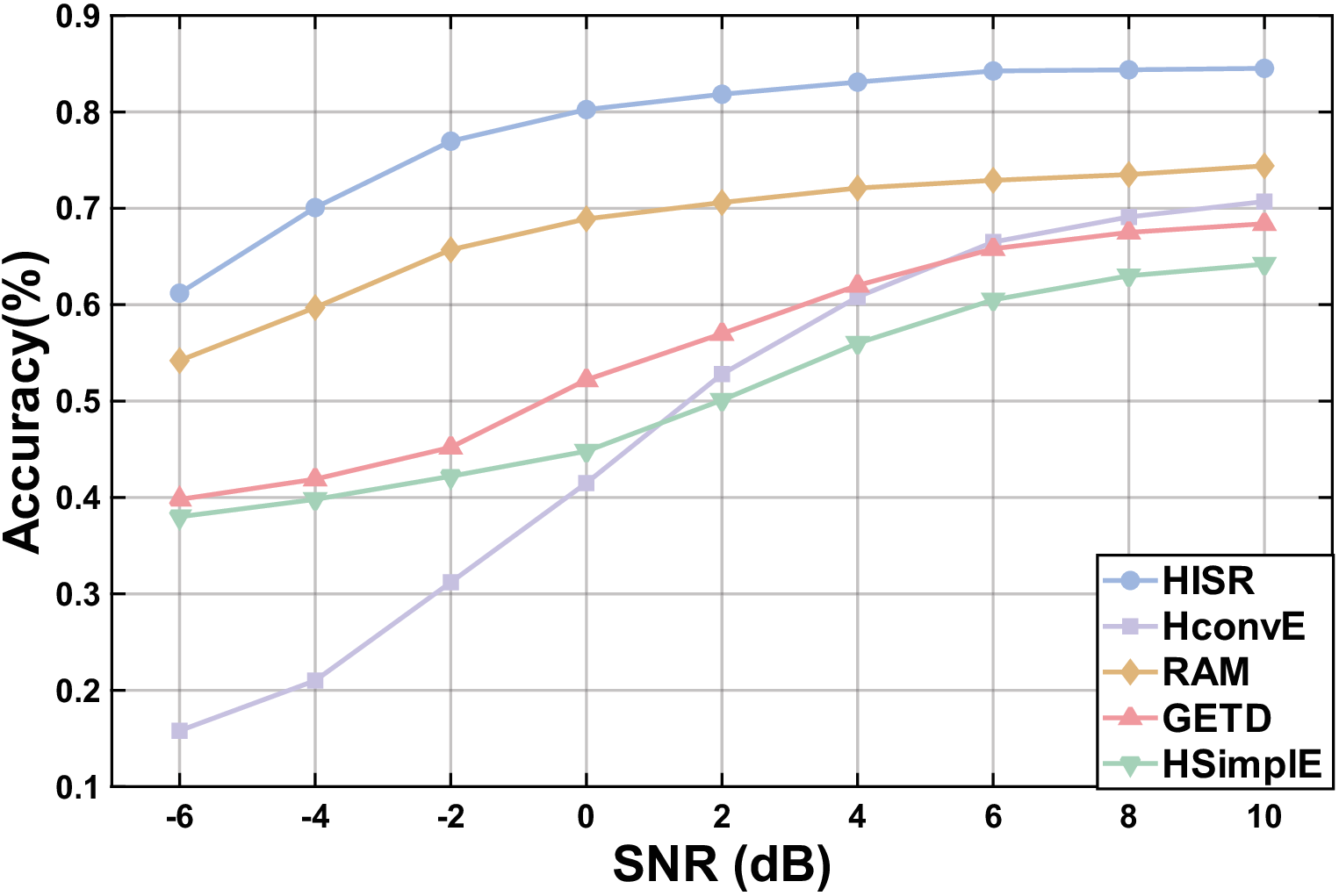}
  \caption{\small{Accuracy of semantic recovery when the channel experiences additive Gaussian noise under different SNRs on the Cora dataset.}}
  \label{Acc-gaussian-cora}
  \end{minipage}
  \hfill
  \begin{minipage}{0.48\linewidth}
   \centering
   \includegraphics[width=4.2cm]{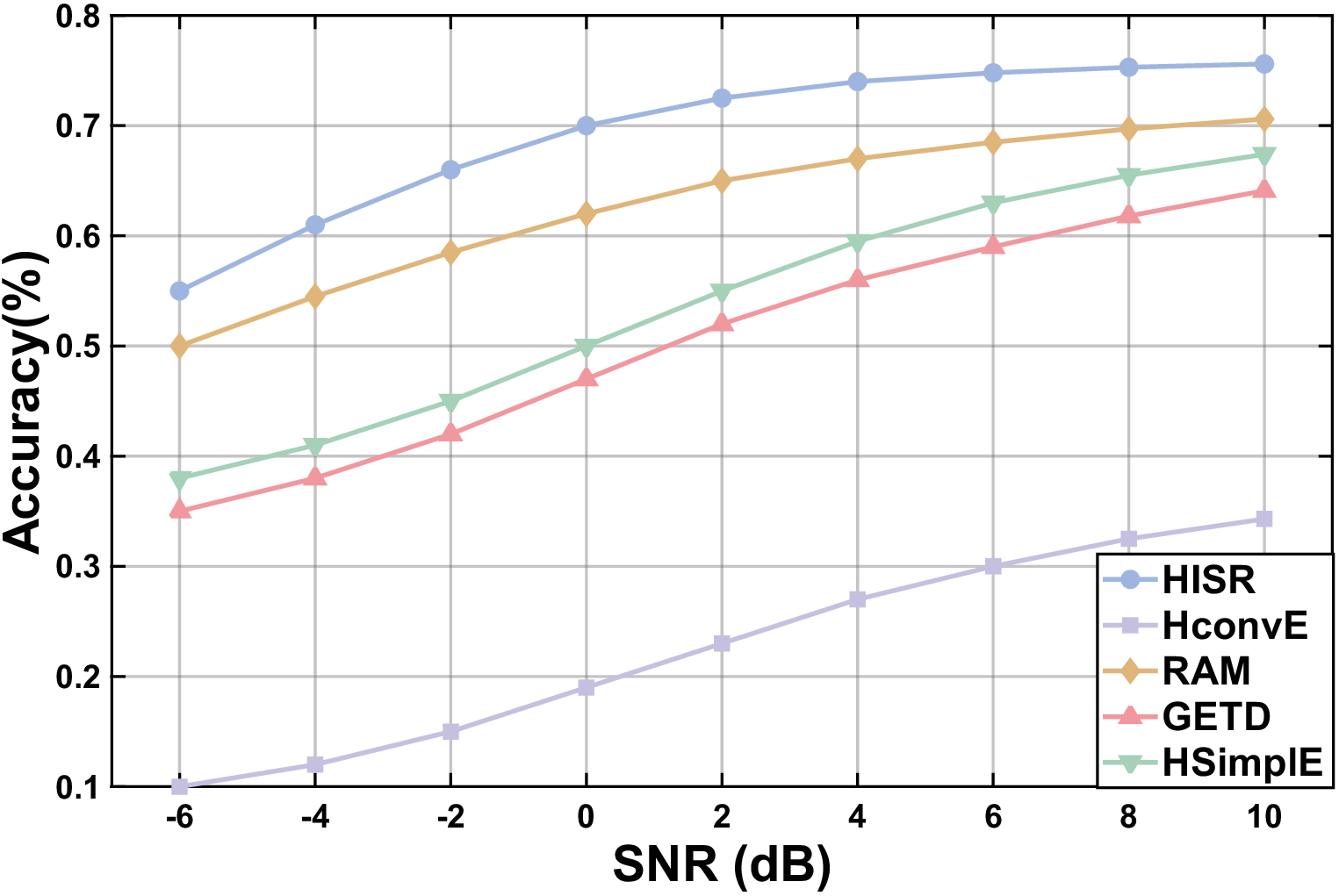}
  \caption{\small{Accuracy of semantic recovery when the channel experiences additive Gaussian noise under different SNRs on the Pubmed dataset.}}
  \label{Acc-gaussian-pubemd}
  \end{minipage}
\end{figure}

Fig.~\ref{Acc-gaussian-cora} and Fig.~\ref{Acc-gaussian-pubemd} compare the semantic recovery accuracy of the proposed HISR framework against several state-of-the-art hypergraph-based baseline methods (HGTN, T-MPNN, HGNN, HyperGCN, CEGCN, and CEGAT) under varying additive Gaussian noise conditions, quantified by signal-to-noise ratio (SNR), on the Cora and Pubmed datasets, respectively. On both datasets, the accuracy generally increases as the SNR improves, indicating better semantic recovery performance under higher-quality transmission conditions. Notably, HISR achieves the highest accuracy of approximately 78.5\% on Cora and 84.9\% on Pubmed at 10dB, better than other baseline methods, demonstrating the robustness and superior noise resilience of our relation-specific semantic subspace embedding and inference mechanisms.

\begin{figure}
  \begin{minipage}{0.48\linewidth}
   \centering
   \includegraphics[width=4.2cm]{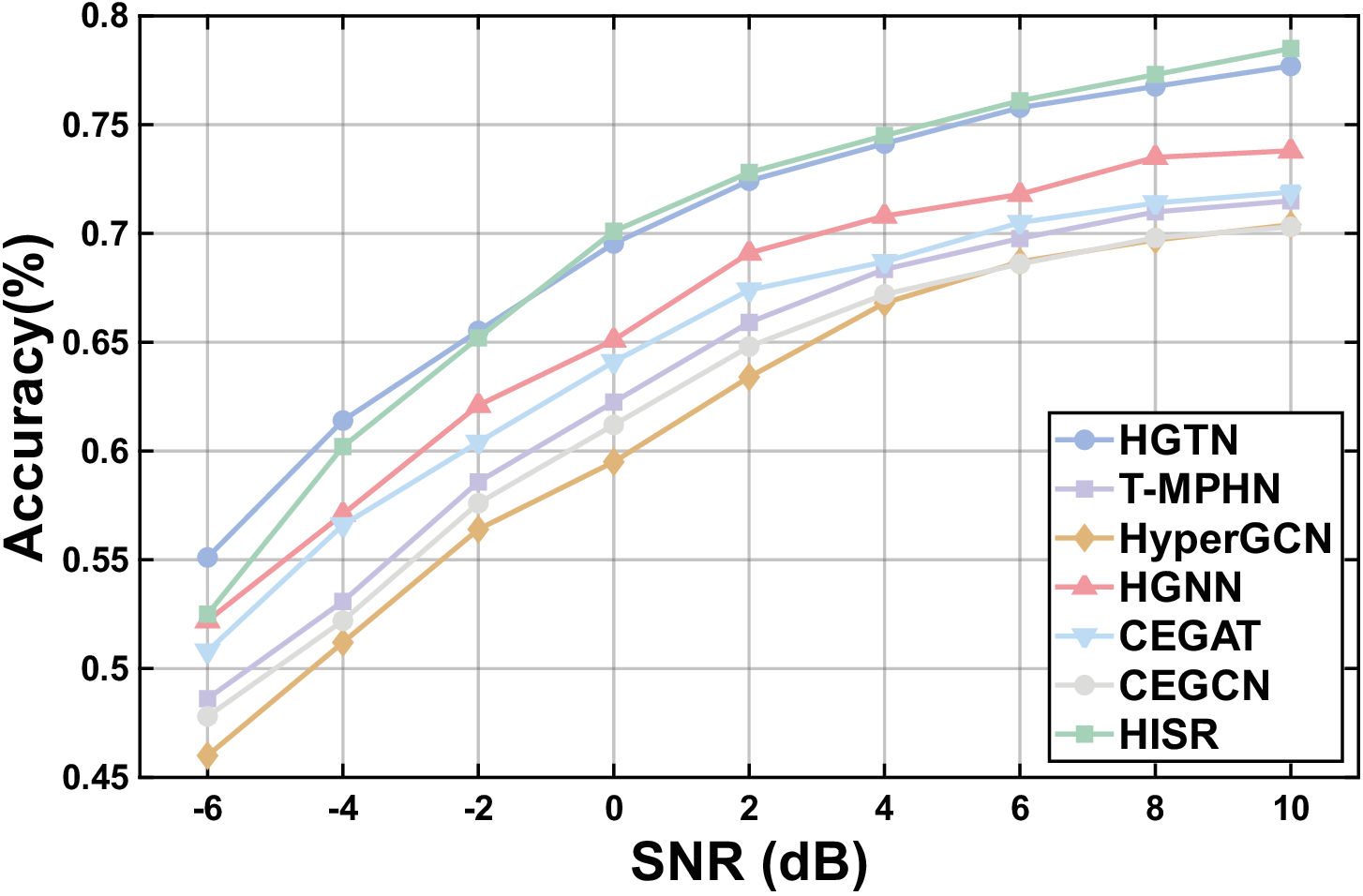}
  \caption{\small{Accuracy of semantic recovery when the channel experiences additive Rayleigh noise under different SNRs on the Cora dataset.}}
  \label{Acc-rayleigh-cora}
  \end{minipage}
  \hfill
  \begin{minipage}{0.48\linewidth}
   \centering
   \includegraphics[width=4.2cm]{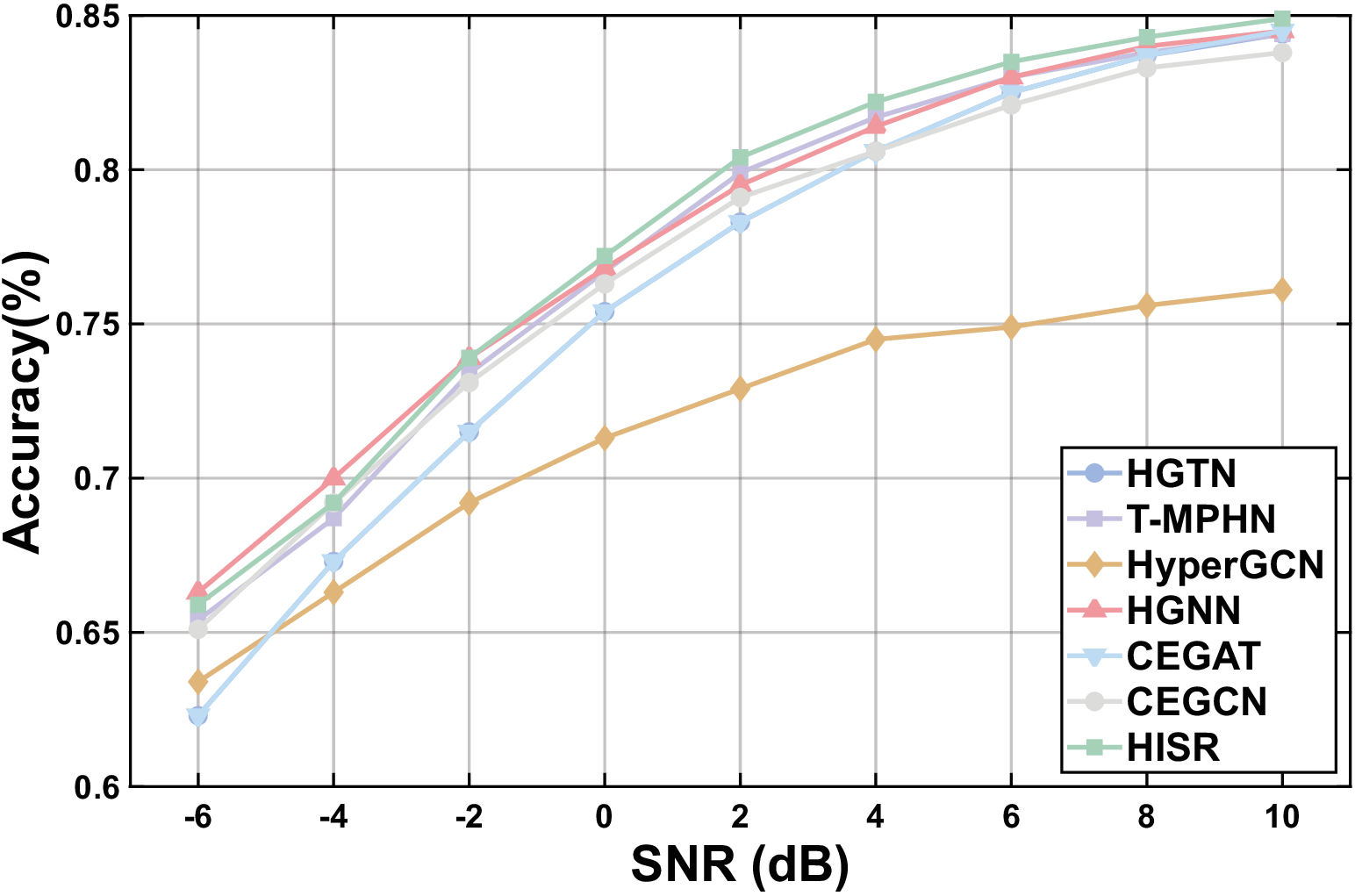}
  \caption{\small{Accuracy of semantic recovery when the channel experiences additive Rayleigh noise under different SNRs on the Pubmed dataset.}}
  \label{Acc-rayleigh-pubemd}
  \end{minipage}
\end{figure}

Fig.~\ref{Acc-rayleigh-cora} and Fig.~\ref{Acc-rayleigh-pubemd} illustrate the accuracy comparisons of semantic recovery under Rayleigh channel conditions across varying signal-to-noise ratios (SNR) for the Cora and Pubmed datasets, respectively. Similar to the additive Gaussian noise scenario, we observe a clear improvement in accuracy as the SNR increases. Notably, HISR maintains superior performance across most of the evaluated SNR levels compared to baseline models (HGTN, T-MPNN, HGNN, HyperGCN, CEGCN, HGTN, and CEGAT), achieving peak accuracies of approximately 74.8\% on Cora and 86.4\% on Pubmed at 10dB. The advantage of HISR is particularly pronounced at lower SNR values, highlighting its robustness and effectiveness in handling severe Rayleigh fading conditions. These results demonstrate that the proposed semantic subspace embedding and inference framework can reliably recover semantic information even under highly challenging and realistic channel environments.

The observed performance saturation at higher SNR levels can be explained as follows. Initially, increasing the SNR significantly reduces communication noise and channel-induced distortion, thereby substantially improving semantic inference accuracy. However, beyond a certain SNR threshold, the inference performance improvement diminishes and eventually saturates. This phenomenon is attributed primarily to the inherent modeling capacity of the semantic embedding and inference framework, as well as residual semantic ambiguity intrinsic to the underlying hypergraph representation. Thus, further improvements in SNR beyond this critical point yield limited performance gains, indicating that performance at higher SNR levels is predominantly constrained by model expressiveness and intrinsic semantic representation quality rather than channel quality itself.

\begin{figure}
  \begin{minipage}{0.48\linewidth}
   \centering
   \includegraphics[width=4.2cm]{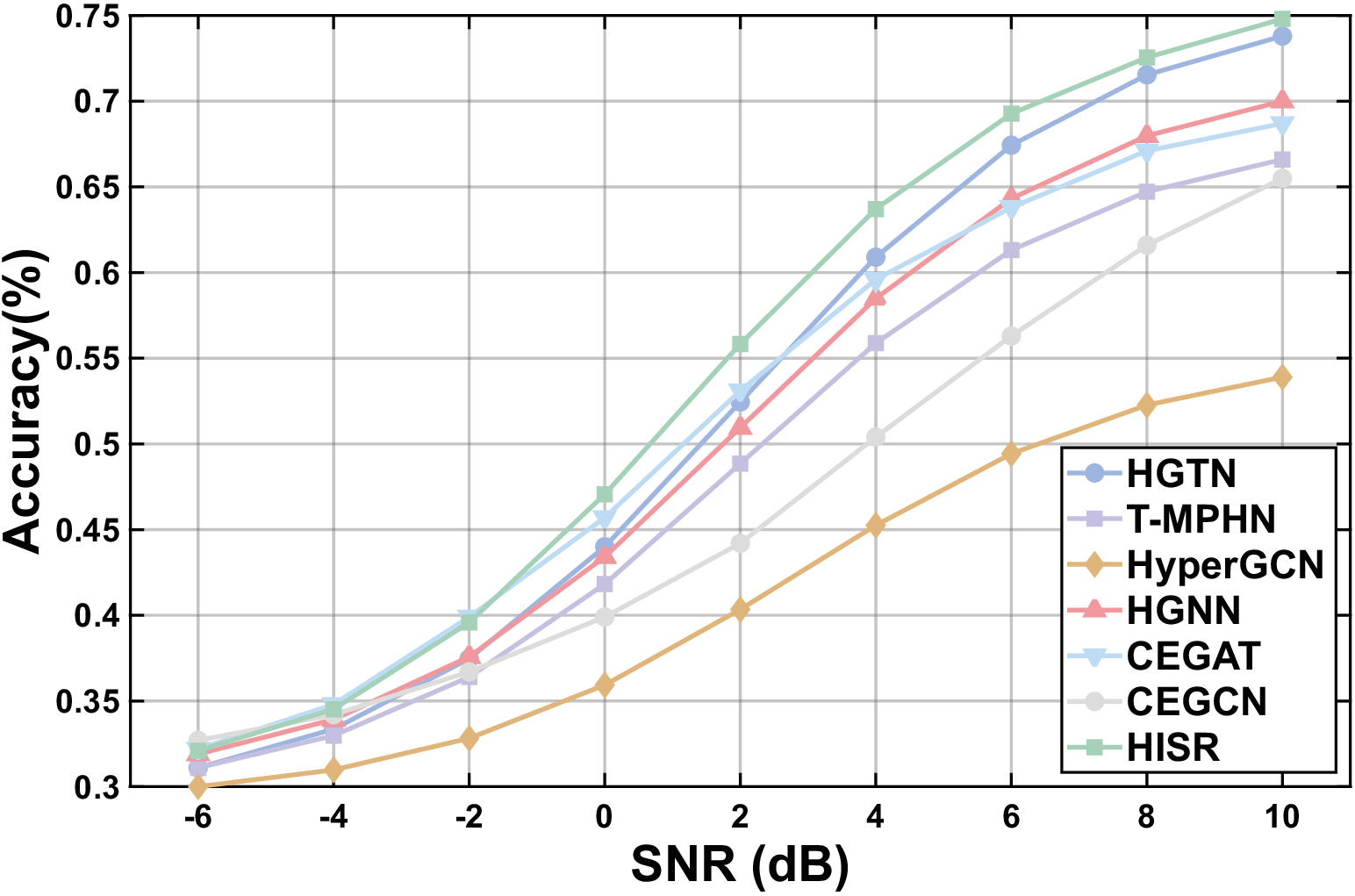}
  \caption{\small{Impact of embedding dimension on classification accuracy on the Cora dataset.}}
  \label{Acc-Dimension-cora}
  \end{minipage}
  \hfill
  \begin{minipage}{0.48\linewidth}
   \centering
   \includegraphics[width=4.2cm]{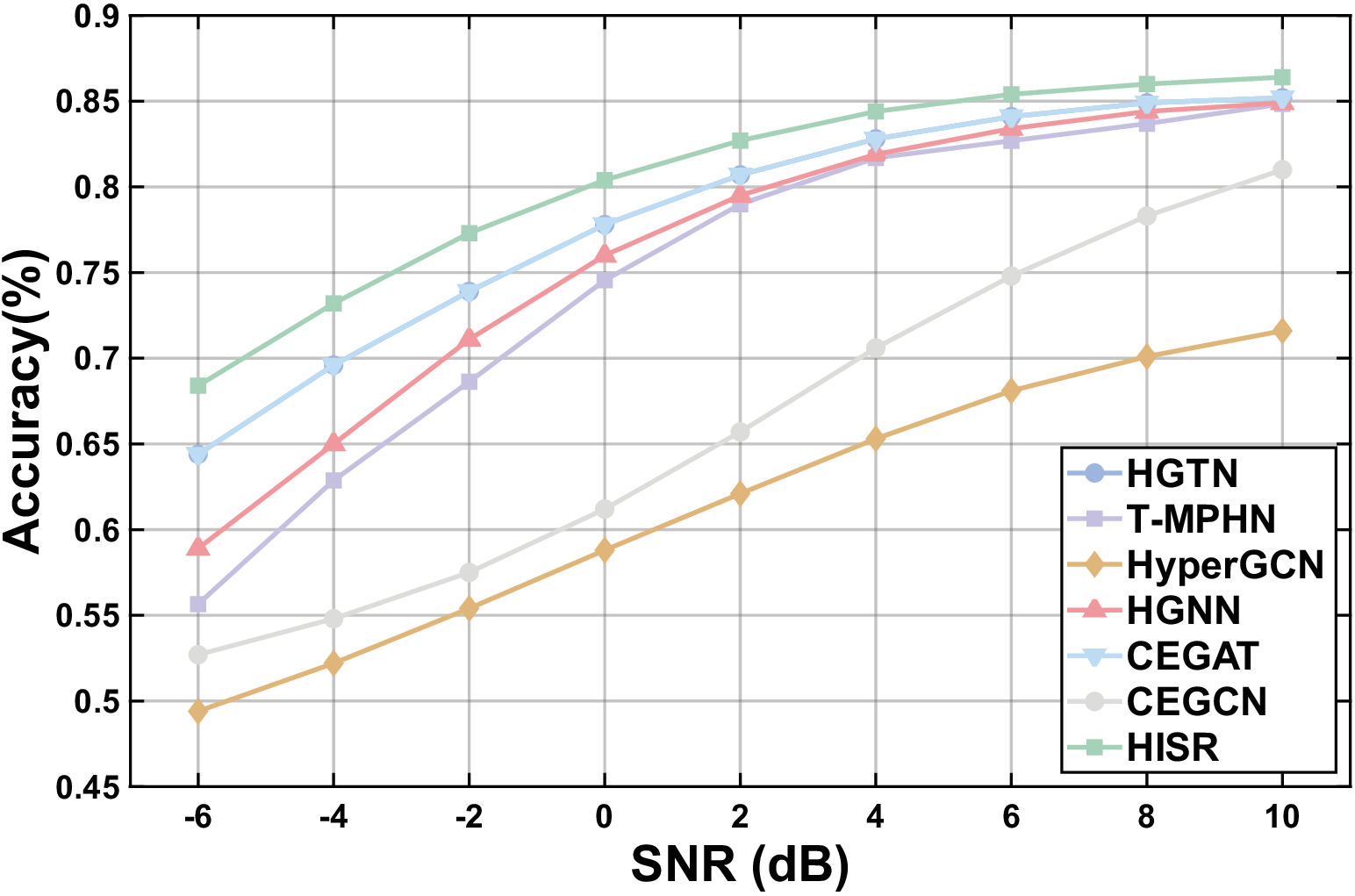}
  \caption{\small{Impact of embedding dimension on classification accuracy on the Pubmed dataset.}}
  \label{Acc-Dimension-pubemd}
  \end{minipage}
\end{figure}

Fig.~\ref{Acc-Dimension-cora} and Fig.~\ref{Acc-Dimension-pubemd} investigate the impact of embedding dimensionality on classification accuracy for the Cora and Pubmed datasets, respectively. For both datasets, classification accuracy rapidly improves as the embedding dimension increases from smaller values, quickly stabilizing at around dimension 15. Notably, HISR consistently achieves the highest accuracy compared to baseline methods (HGTN, T-MPNN, HGNN, HyperGCN, CEGCN, and CEGAT) across all embedding dimensions, starting at approximately 57.0\% and 60.6\% at dimension 2, and reaching peaks of 79.2\% and 73.9\% for Cora and Pubmed. This observation underscores the effectiveness of our semantic subspace embedding approach, which efficiently captures richer semantic information even at relatively low embedding dimensions, thereby demonstrating superior representation power and scalability.


\section{Conclusion}

In this paper, we have introduced a novel hypergraph-based semantic communication framework, termed HISR, designed specifically to address the limitations of traditional pairwise graph-based semantic embedding methods. By explicitly modeling and reasoning about higher-order semantic relationships through dedicated semantic subspaces, HISR provides a robust and efficient solution to encode, transmit, and reconstruct complex multi-entity semantic information. The proposed semantic subspace construction mechanism effectively mitigates the prevalent over-smoothing issue by preserving distinct semantic identities within relation-specific subspaces, significantly enhancing the clarity and reliability of semantic embeddings.

We further presented an implicit semantic inference strategy that leverages these semantic subspaces to accurately recover missing or corrupted semantic entities, employing a position-aware aggregation and scoring mechanism. This ensures robust and precise semantic reasoning even under conditions of noisy communication channels and incomplete information scenarios.

Comprehensive theoretical analyses were provided to rigorously justify the effectiveness and efficiency of our proposed approach, alongside extensive experimental validation across multiple benchmark datasets. Experimental results consistently demonstrated that HISR substantially outperforms existing state-of-the-art methods in terms of semantic reconstruction accuracy, robustness to noise, and computational efficiency.

Overall, our work establishes a strong foundation for future research on hypergraph-based semantic communication systems, offering practical insights and a powerful methodological framework for achieving expressive, reliable, and efficient semantic understanding in next-generation communication networks.

\bibliography{reference}
\bibliographystyle{IEEEtran}

\begin{IEEEbiography}[{\includegraphics[width=1.1in,height=1.3in,clip,keepaspectratio]{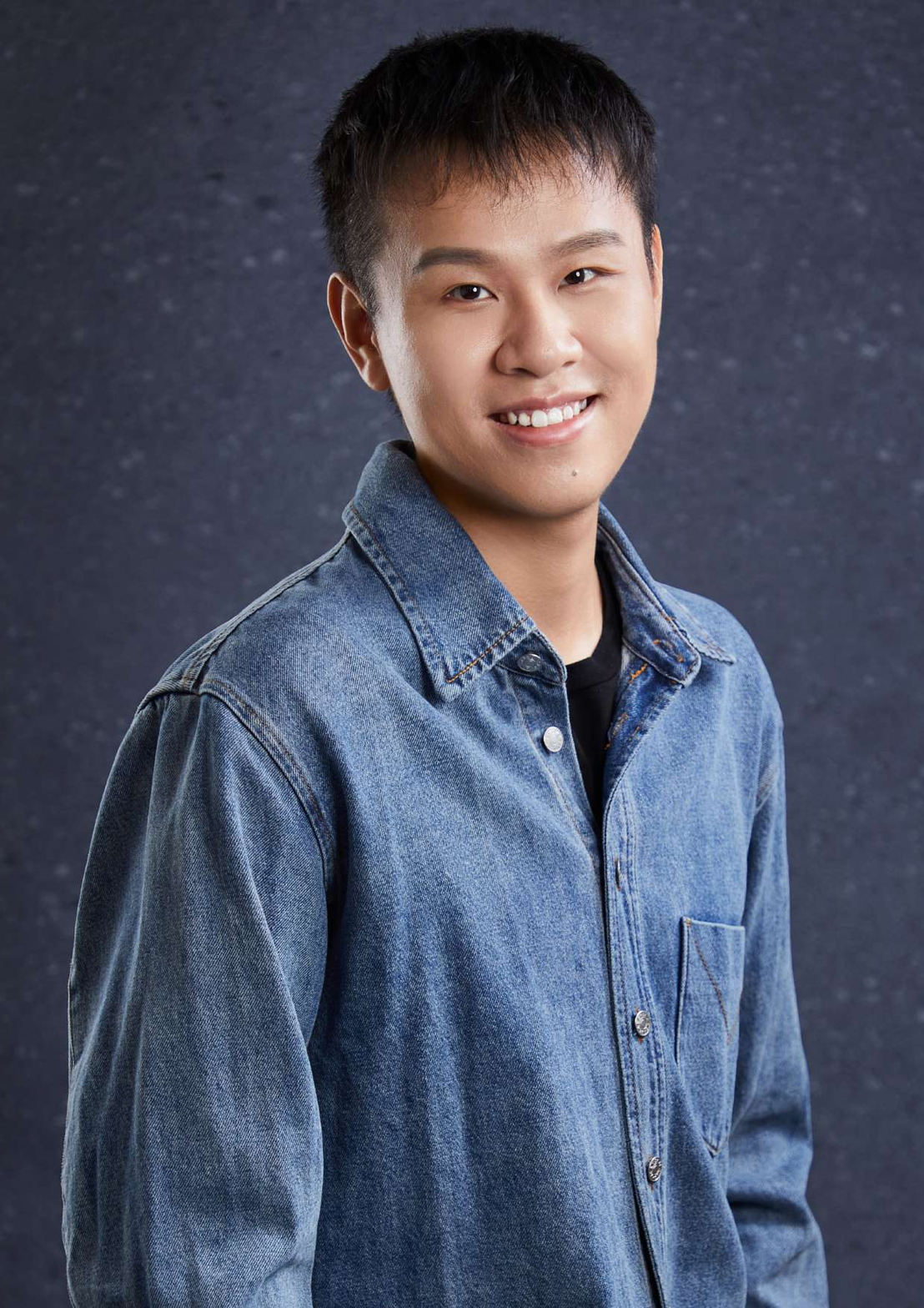}}]{Yiwei Liao}  received his B.S. degree from the School of Electrical and Electronic engineering from Huazhong University of Science and Technology, Wuhan, China in 2017, and M.S. degree in Northeastern University, MA, US in 2020 and Ph.D degree in the school of electronic information and communications at the Huazhong University of Science and Technology in 2025, Wuhan, China. He is now a engineer in China Electric Power Research Institute, Wuhan, China. His research interest includes network AI and next generation communication technology, Intelligent inspection of power equipment by drone. 
\end{IEEEbiography}

\begin{IEEEbiography}[{\includegraphics[width=1.1in,height=1.3in,clip,keepaspectratio]{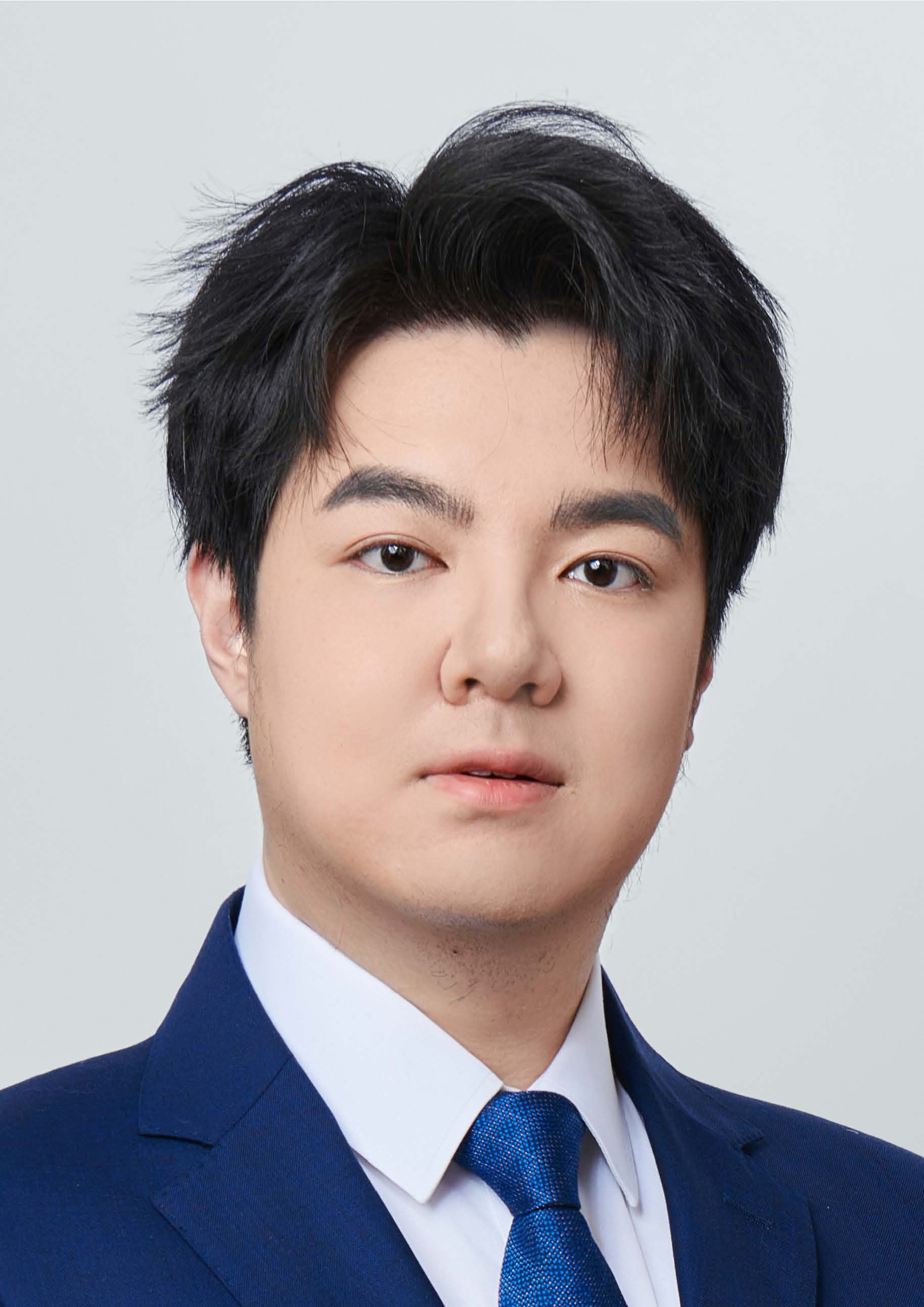}}]{Shurui Tu}  received the B.S. degree from the School of Optical and Electronic Information, Huazhong University of Science and Technology, Wuhan, China, in 2024. He is currently working toward the M.S. degree with the School of Electronic Information and Communications, Huazhong University of Science and Technology. His research interests center on semantic communications and knowledge hypergraphs, focusing on knowledge reasoning in wireless systems.
\end{IEEEbiography}

\begin{IEEEbiography}[{\includegraphics[width=1.1in,height=1.3in,clip,keepaspectratio]{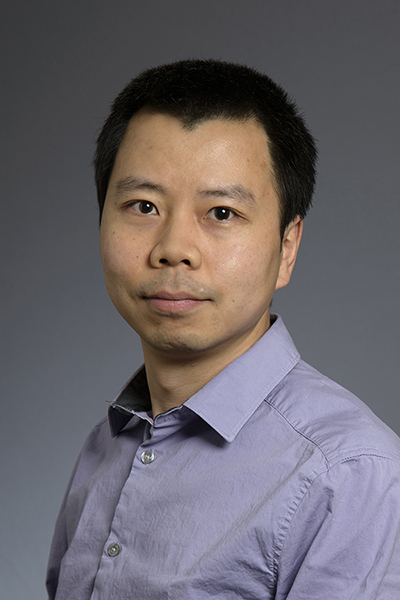}}]{Yong Xiao} (Senior Member, IEEE) received his B.S. degree in electrical engineering from China University of Geosciences, Wuhan, China in 2002, M.Sc. degree in telecommunication from Hong Kong University of Science and Technology in 2006, and his Ph. D degree in electrical and electronic engineering from Nanyang Technological University, Singapore in 2012. He is now a professor in the School of Electronic Information and Communications at the Huazhong University of Science and Technology (HUST), Wuhan, China. He is also with Peng Cheng Laboratory, Shenzhen, China, and Pazhou Laboratory (Huangpu), Guangzhou, China. He currently serves as the associate group leader of the network intelligence group for IMT-2030 (the 6G promoting group). He leads the first international standardization effort on semantic-aware networking at ITU-T. Before he joined HUST, he was a research assistant professor in the Department of Electrical and Computer Engineering at the University of Arizona, where he served as the center manager of the Broadband Wireless Access and Applications Center (BWAC), an NSF Industry/University Cooperative Research Center (I/UCRC) led by the University of Arizona. His research interests include AI/ML, game theory, distributed optimization, and their applications to agentic AI networking, semantic-aware networking, and semantic communications.
\end{IEEEbiography}

\begin{IEEEbiography}[{\includegraphics[width=1.1in,height=1.3in,clip,keepaspectratio]{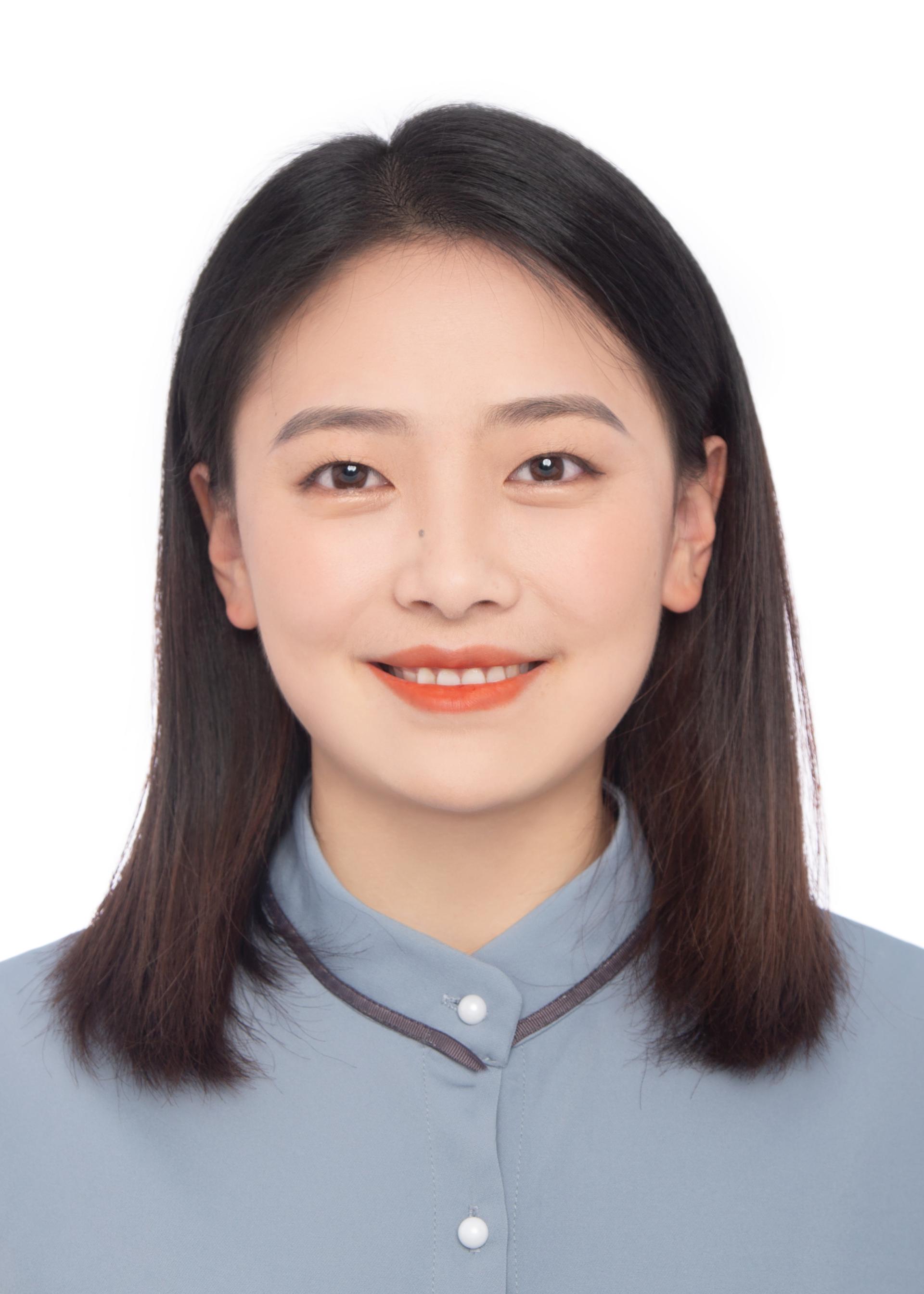}}]{Yingyu Li} (Member, IEEE) received the B.Eng. degree in electronic information engineering and the Ph.D. degree in circuits and systems from Xidian University, Xi'an, China, in June 2012 and September 2018, respectively. From September 2014 to September 2016, she was a Research Scholar with the Wireless Networking, Signal Processing and Security Lab, Department of Electronic Computer Engineering, University of Houston, USA. She was a postdoctoral researcher in the School of Electronic Information and Communications at Huazhong University of Science and Technology from October 2018 to November 2021. Since December 2021, she has been 
is an Associate Professor at the School of Mechanical Engineering and Electronic Information, China University of Geosciences (Wuhan). Her research interests include machine learning and artificial intelligence for next-generation wireless networks, federated edge intelligence, green/low-carbon communication networks, distributed optimization, semantic communications, and intelligent Internet of Things.
\end{IEEEbiography}

\begin{IEEEbiography}[{\includegraphics[width=1.1in,height=1.3in, clip,keepaspectratio]{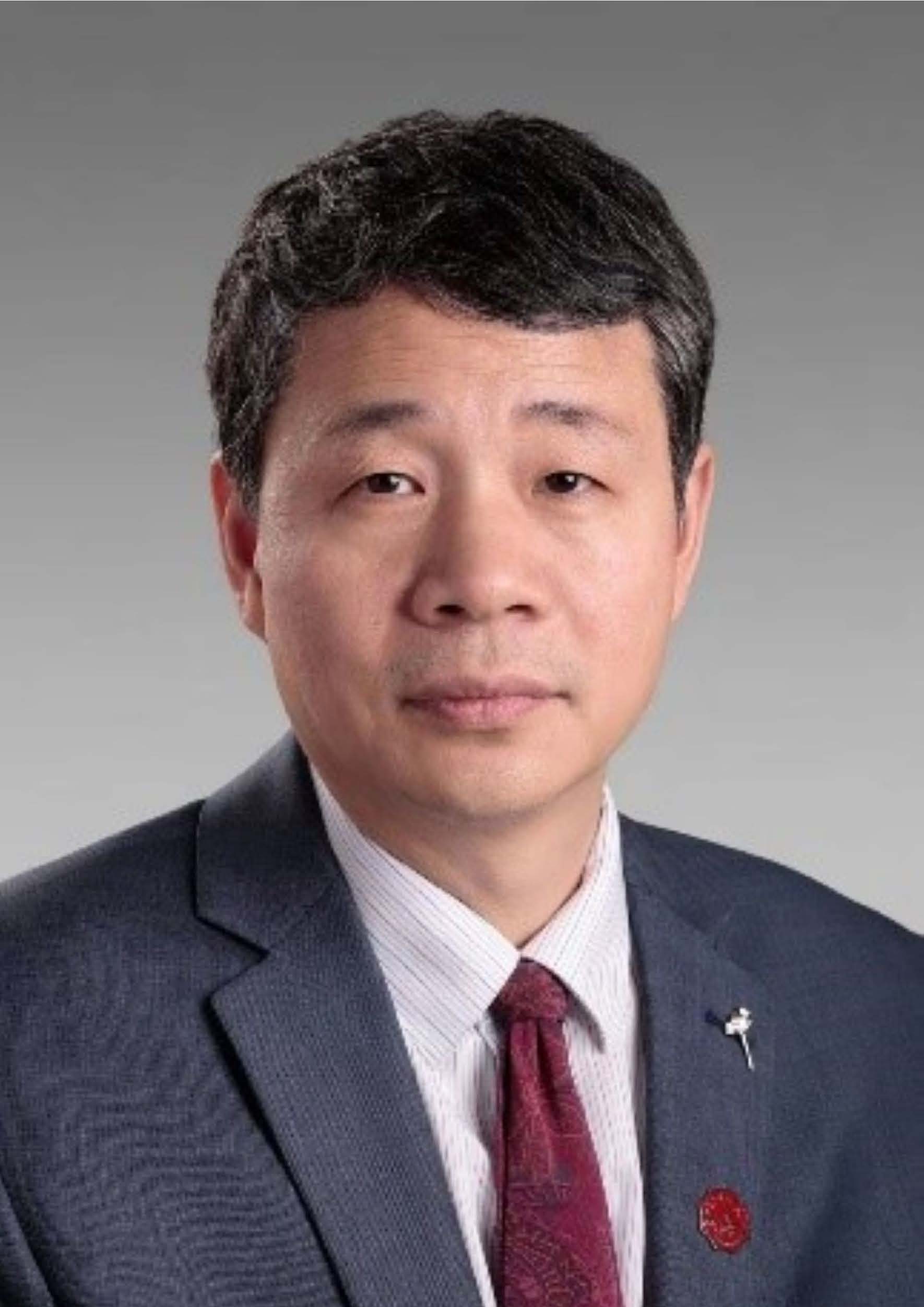}}]{Guangming Shi} (Fellow, IEEE) received the M.S. degree in computer control, and the Ph.D. degree in electronic information technology from Xidian University, Xi’an, China, in 1988, and 2002, respectively. He was the vice president of Xidian University from 2018 to 2022. Currently, he 
is the Vice Dean of Peng Cheng Laboratory and a Professor with the School of Artificial Intelligence, Xidian University. He is an IEEE Fellow, the chair of IEEE CASS Xi’an Chapter, a senior member of ACM and CCF, Fellow of the Chinese Institute of Electronics, and Fellow of IET. 
He won the second prize of the National Natural Science Award in 2017. His research interests include Artificial Intelligence, Semantic Communications, and Human-Computer Interaction.
\end{IEEEbiography}

\end{document}